\documentclass[11pt,a4paper]{article}
\usepackage{acl2023}
\usepackage{times}
\usepackage{latexsym}
\usepackage{amsmath}
\usepackage{color}
\usepackage{booktabs}
\usepackage{graphics}
\usepackage{graphicx}
\usepackage{amsfonts}
\usepackage{url}
\usepackage[T1]{fontenc}
\usepackage{inconsolata}
\usepackage{xspace}
\usepackage{enumitem}
\usepackage{subcaption}
\usepackage{multirow}
\usepackage{microtype}

\newcommand{\secref}[1]{\S\ref{#1}}
\newcommand{\model}{{\tt \textbf{MaNtLE}}\xspace}
\newcommand{\lime}{{{LIME}}\xspace}

\newcommand{\clues}{{\tt {CLUES}}\xspace}
\newcommand{\cluesreal}{{\tt \clues-Real}\xspace}

\newcommand{\quexent}{{\tt {LaSQuE}}\xspace}
\newcommand{\anchors}{{{Anchors}}\xspace}
\newcommand{\Beam}{{\tt {Beam}}\xspace}
\newcommand{\Perfeat}{{\tt {PerFeat}}\xspace}

\definecolor{LightCyan}{RGB}{172,204,186}

\title{\model: \underline{M}odel-\underline{a}gnostic \underline{N}a\underline{t}ural \underline{L}anguage \underline{E}xplainer}

\author{Rakesh R Menon \qquad Kerem Zaman \qquad Shashank Srivastava \\
    UNC Chapel Hill \\
  \texttt{\{rrmenon, kzaman, ssrivastava\}@cs.unc.edu} \\}

\begin{document}
\maketitle

\begin{abstract}
    Understanding the internal reasoning behind the predictions of machine learning systems is increasingly vital, given their rising adoption and acceptance. While previous approaches, such as \lime, generate algorithmic explanations by attributing importance to input features for individual examples, recent research indicates that practitioners prefer examining \emph{language explanations that explain sub-groups of examples} \cite{lakkaraju2022rethinking}. In this paper, we introduce \model, a model-agnostic natural language explainer that analyzes multiple classifier predictions and generates \emph{faithful natural language explanations} of classifier rationale for structured classification tasks. \model uses multi-task training on thousands of synthetic classification tasks to generate faithful explanations. Simulated user studies indicate that, on average, \model-generated explanations are at least 11\% more faithful compared to \lime and \anchors explanations across three tasks. Human evaluations demonstrate that users can better predict model behavior using explanations from \model compared to other techniques.\footnote{Work in progress}
\end{abstract}
\section{Introduction} 
\label{sec:intro}
% What is the current situation, and what is the problem?
The increasing adoption of black-box machine learning models across various applications~\cite{credit-risk, DADA2019e01802} has led to a pressing need for generating human-understandable explanations of their decision-making process. 
While such models may yield high predictive accuracies, their underlying reasoning often remains opaque to their end users. This lack of transparency is a critical barrier to their adoption in critical domains, such as healthcare, law, and medicine. 

% What has been tried, and what are the shortcomings?
To interpret decisions made by machine learning models, prior work has proposed multiple techniques including feature importances (\lime, \citet{ribeiro2016should}), rule lists (\anchors, \citet{ribeiro2018anchors}), and model-generated explanations \citep{rajani-etal-2019-explain, narang2020wt5}. 
However, these explanations predict model behavior at the level of individual examples rather than subgroups, which makes it hard for designers to identify systemic problems and suggest model improvements. Recent work suggests that this shortcoming contributes to practitioners' reluctance to utilize machine learning systems for critical applications~\cite{lakkaraju2022rethinking}.

\begin{figure}[!t]
    \centering
    \includegraphics[width=0.47\textwidth]{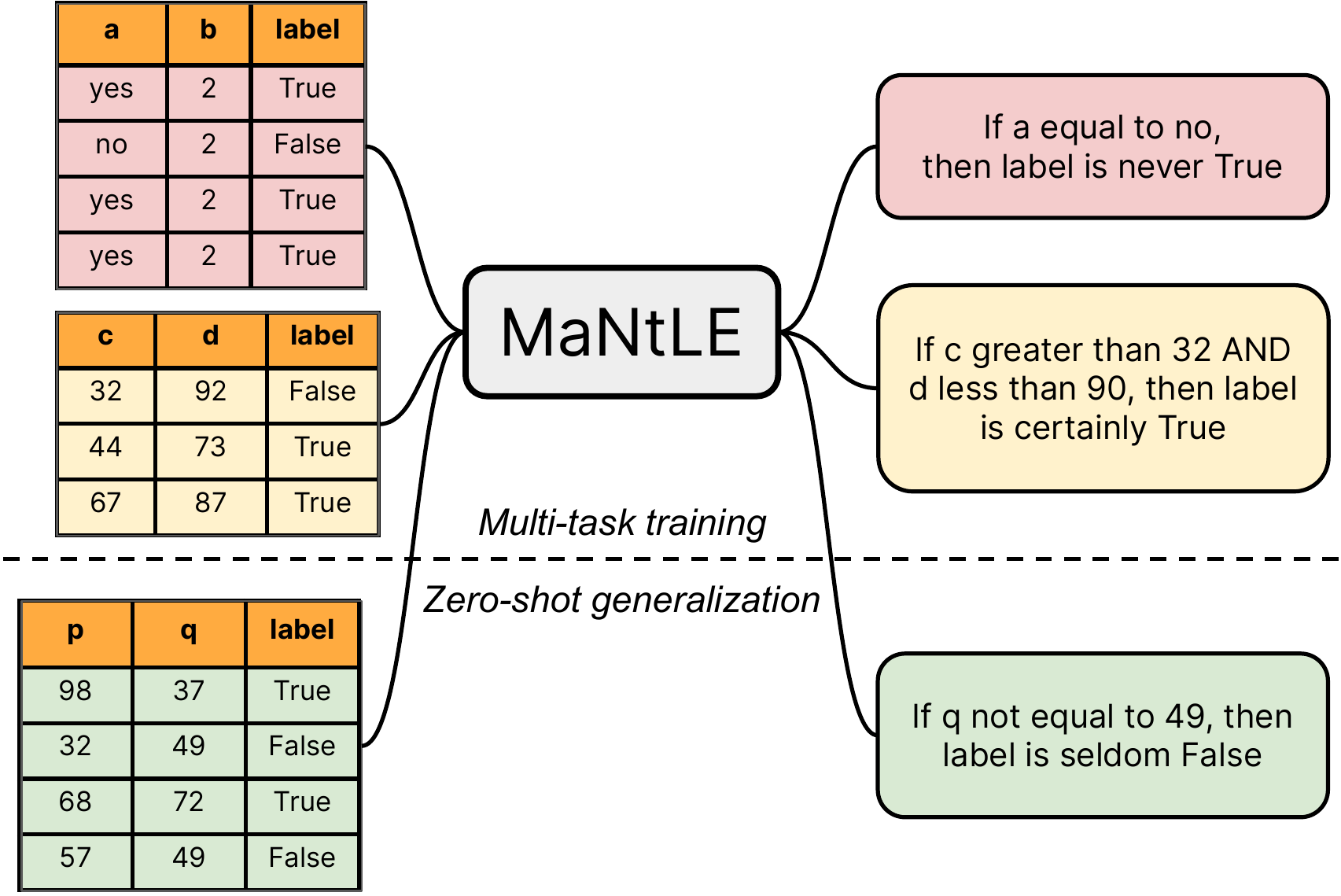}
    \caption{\model is a model-agnostic natural language explainer. Following massive multi-task learning, \model can generate explanations of decision-making rationale for new classifiers and datasets. 
    }
    \label{fig:multitask_mantle}
\end{figure}

% What is the proposed solution?
In this paper, we introduce \model, a model-agnostic natural language explainer that analyzes multiple classifier predictions and generates \emph{faithful natural language explanations} of the classifier's reasoning for structured classification tasks, as depicted in Figure~\ref{fig:multitask_mantle}. The goal of \model is to explain the rationale of classifiers on real-world tasks. To develop \model, we fine-tune a {\tt T5-Large} model on thousands of synthetic classification tasks, each paired with natural language explanations, in a multi-task learning setup following recent research~\cite{wei2022finetuned, sanh2022multitask, mishra-etal-2022-cross, menon-etal-2022-clues}. In \secref{sec:inf_tech}, we discuss inference procedures to improve explanation quality and adapt the model trained on synthetic data for real-world tasks.

% How well does it address the problems? What is the proof of the solution working?
We test \model explanations on real-world tasks by assessing their ability to aid explanation-guided zero-shot classifiers in the \cluesreal benchmark \cite{menon-etal-2022-clues}. 
Our results show that \model explanations are as helpful as human-written explanations in guiding classifiers (\secref{sec:interpret_datasets}).
Next, we compare the faithfulness \cite{jacovi-goldberg-2020-towards} and simulatability \cite{hase-bansal-2020-evaluating} of explanations generated by \model, \lime, and \anchors for four classification techniques on three real-world tasks (\secref{sec:simuser_exp}). 
In budget-constrained scenarios, where the number of available examples is comparable across methods, \model explanations are $37\%$ and $11\%$ more faithful on average than \lime and \anchors, respectively.

In user studies (\secref{sec:human_exp}), we observe that users without strict machine learning expertise prefer explanations from \model over attribution-score-based explanations by \lime (overall preference of $3.44$ vs $2.16$ on a five-point Likert scale; $p<0.001$ t-test). 
Further, users can predict model behavior better using \model explanations in at least $25\%$ more instances than \lime and \anchors for the \texttt{adult} dataset \citep{Dua:2019}.
Our results corroborate the findings of \citet{lakkaraju2022rethinking} on the need for providing natural language explanations that explain subgroups of examples to help practitioners interpret ML models.

Finally, we analyze the contributions of
the multitask-training dataset and model size to \model's generated explanations.
Further, we show that increasing the number of examples accessible by \model can enhance explanation quality, 
highlighting possibilities for using the model in resource-rich settings. 
\\
% What is the summary of the contributions made by this work?
In summary, our contributions are:
\begin{itemize}[itemindent=-0em, labelsep=0.1cm, leftmargin=0.6em, itemsep=0.05em]
    \vspace{-0.1cm}
    \item We develop \model, and 
    demonstrate its efficacy on real-world tasks by comparing (a) classification-utility of explanations with human-written explanations on the \cluesreal benchmark, (b) faithfulness and simulatability of explanations with popular approaches. 
    \item We show that users predict model behavior better with \model explanations compared to \lime and \anchors. Users also rate \model explanations as better in understandability, informativeness, and overall preference.
    \item We analyze factors contributing to \model's performance and outline opportunities for improving explanations.
    
\end{itemize} 
\section{Related Work}

\paragraph{Explainability methods.} 

Previous research has proposed methods for understanding model rationale, which can be broadly categorized into feature attribution and language explanation methods. 

Feature attribution techniques \cite{pmlr-v80-kim18d, lundberg2017unified, pmlr-v70-sundararajan17a} provide information about how models use certain features to derive classification decisions, utilizing methods such as sparse linear regression models \citep[\lime]{ribeiro2016should} and rule lists \citep[\anchors]{ribeiro2018anchors}. 
These methods have two shortcomings: (a) they explain model behavior in a local region around an example, and (b) they generate explanations by solving a search problem over a set of instances from the original training data or by generating perturbations around an example (typically, $\sim 1000$ examples).
However, gaining access to model predictions on thousands of examples can be resource-intensive or financially unviable for large models, such as GPT-3 \cite{NEURIPS2020_1457c0d6}.
In contrast, we introduce a method that looks at tens of examples and generates explanations without additional data or perturbations.
Some recent works such as CAGE \cite{rajani-etal-2019-explain} and WT5 \cite{narang2020wt5} have explored natural language explanations for reasoning and rationalizing classification decisions for language understanding tasks.
Like \lime and \anchors, these explanations are specific to individual examples. 
Crucially, the function of explanations in these works is to help improve classification performance rather than to interpret model behavior. 
Further, training these models requires thousands of human-written explanations for a task of interest which is often impractical.
Our work is distinct from this line of research as we seek to explain model behavior rather than improving classification models, using only a few examples. 

\paragraph{Multi-task Training of Language Models.} 
Large language models (LLMs) pretrained on large text corpora have shown impressive performances on various tasks \cite{NEURIPS2020_1457c0d6, scao2022bloom}.
Some recent works \cite{wei2022finetuned, sanh2022multitask} have explored multitask training of language models on labeled datasets paired with natural language instructions to enhance zero-shot generalization, a procedure called instruction fine-tuning. 
In contrast, \model generates explanations when prompted with feature-prediction pairs.
\section{\model} 
\label{sec:method}

\subsection{Problem Setup}
We assume access to a set of input-prediction pairs, $\{(X_i, y_{\theta_t,i})_{i:1\rightarrow N}\}$, from a black-box classifier, $\theta_t$, trained for a structured classification task $t$. 
Formally, we define the model (or classifier) rationale
as a natural language statement that describes the feature constraints 
used by the classifier to predict $y_{\theta_t,1:M}$ for $M$ inputs $X_{1:M}$ (see Figure~\ref{fig:multitask_mantle}). 
The aim of the explainer, here \model, is to derive explanations, $e \in \mathcal{E}$, 
where $\mathcal{E}$ is the set of all possible natural language explanations.

\subsection{Model} \label{sec:model}
\begin{figure}[b]
    \centering
    \includegraphics[width=0.48\textwidth]{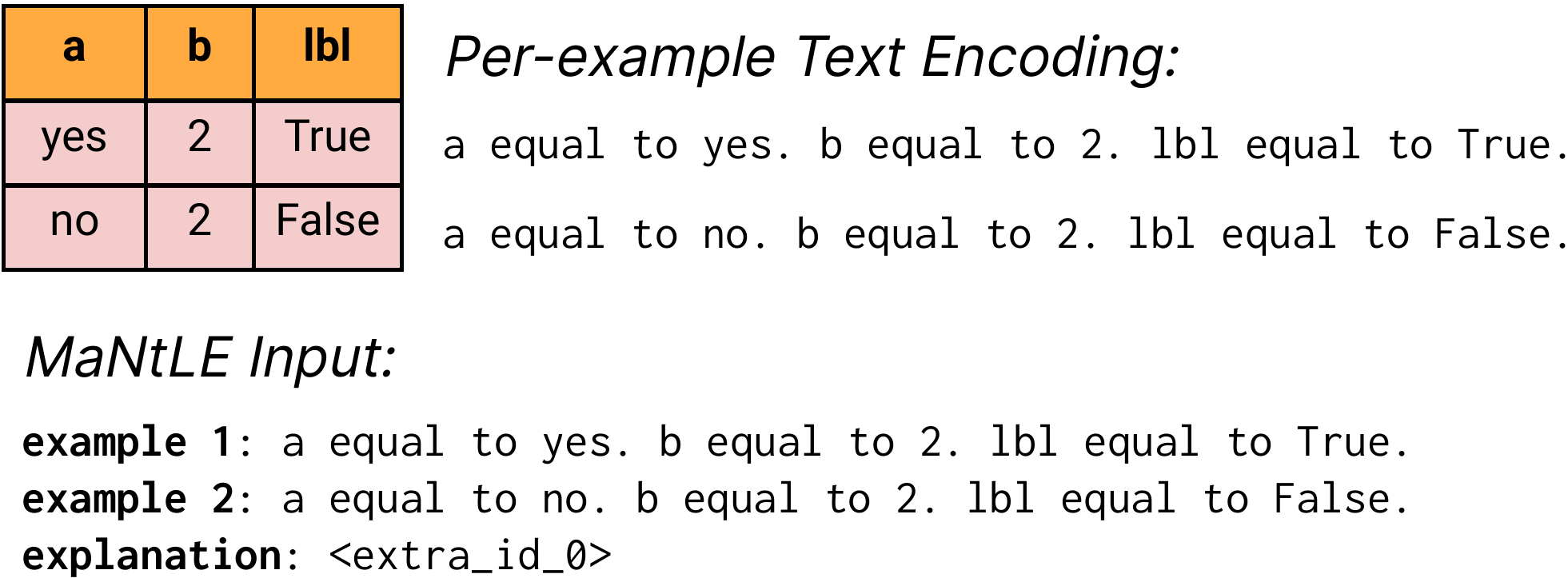}
    \caption{\model's strategy for linearizing structured data to text.
    } 
    \label{fig:text_encoding}
\end{figure}

We frame the task of explanation generation from input-prediction pairs as a sequence-to-sequence problem. 
Specifically, we linearize input-prediction pairs as text and predict text explanations using our model.
Framing the task as a sequence-to-sequence problem enables 
fine-tuning of pre-trained language models to generate language explanations.
We initialize \model using \texttt{T5-Large} \cite{raffel2020exploring} \footnote{Due to computational constraints, we do not experiment with larger models (\texttt{T5-3B} or \texttt{T5-11B}).}
language model.
Our input-linearization strategy converts a set of \textit{k} examples into text with an additional prompt, \texttt{explanation: <extra\_id\_0>}. 
Figure \ref{fig:text_encoding} illustrates the encoding process with two examples.
At the decoder side, we begin prediction from the \texttt{<extra\_id\_0>} sentinel token 
to match the span-prediction objective of \texttt{T5}.

\subsection{Multi-task Training} \label{sec:training_details}

To train \model, we use massive multi-task training following recent advancements in this area \cite{wei2022finetuned, sanh2022multitask, mishra-etal-2022-cross}.
This approach equips \model with the ability to generate explanations for any classifier without the need for additional fine-tuning.
The degree of generalization achieved by the model depends on the number and diversity of tasks used during multi-task training.
However, obtaining language explanations for thousands of classifiers for training purposes is infeasible.
To address this, we adopt a strategy proposed by \citet{menon-etal-2022-clues} and generate synthetic classification tasks programmatically, where the tasks have known natural language explanations and the examples are used to predict these explanations. 
To diversify the set of tasks, we generate tasks with explanations that include a variety of logical operations in addition to explanations conditioned on a single feature, e.g., \textit{`If fruit color is red, then apple'}. 
According to \citet{menon-etal-2022-clues}, the synthetic tasks and explanations vary in terms of the logical operations present, including \textit{conjunctions}, \textit{disjunctions}, \textit{negations}, and \textit{quantifiers}.
We refer the reader to \secref{sec:pretrain_data} and Table \ref{tab:syn-template} for information on the task variations/complexities. 
These variations are based on prior research, which explores the linguistic elements used by humans to communicate task intents through language \cite{Chopra2019TheFC}.

Additionally, we assume that users are interested in understanding the rationale behind a specific class of a classifier at any given point. Thus, we re-frame all examples as binary classification examples for \model, where the label of interest is maintained, as \texttt{\{label\}} say, and the remaining labels are re-labeled as \texttt{``not \{label\}"}. 

\subsection{Inference Techniques}
\label{sec:inf_tech}

For inference, we experiment with three decoding approaches. 
The first is greedy decoding, where we generate the most likely explanation from \model conditioned on the input.
The second approach aims to improve explanations by sampling multiple candidates and returning the explanation that most faithfully explains model behavior on the inputs (see \secref{sec:eval_metrics} for the definition of faithfulness). 
For this, we use beam-search, with beam size 20, and generate 20 explanations from \model, following which we return the most faithful explanation. 
Assessing beam-search generations, we found that sequences often repeated the same feature. To diversify explanations, we develop \Perfeat decoding, our third decoding approach. Here,
we prompt the \model decoder with `\texttt{If \{feature\_name\}}' to generate sequences for each feature and return the most faithful explanation.

\subsection{Training Details}
We perform multi-task training for \model on $\sim200$K synthetic datasets that have five features in all tasks covering a wide range of explanation types (\secref{sec:model}). 
We cap the maximum number of tokens to 1024 and 64 tokens, respectively, for the input and the output. 
This corresponds to packing between 10-12 examples in the input when generating explanations for classifier rationale.
Additionally, since \model derives explanations by extracting patterns in examples from both classes (\texttt{\{label\}} and \texttt{not \{label\}}), we ensure that at least 10\% of the input examples belong to each of the two classes.
The model is trained for 2 million steps with a batch size of 4 using AdamW \cite{loshchilov2018decoupled}, a learning rate of 1e-5, and a linear decay schedule with a warmup of $50$K steps.

After training, we select the best model checkpoint using the highest score (sum of all metrics in \secref{sec:eval_metrics}) on the validation sets of 20 randomly selected synthetic tasks from those used during training.

\subsection{Evaluation Metrics} 
\label{sec:eval_metrics}
To evaluate the generated language explanations, we 
use 
BERTScore \cite{Zhang*2020BERTScore:}, ROUGE \cite{lin-2004-rouge}, and BLEU \cite{10.3115/1073083.1073135}. 
However, these metrics capture surface-level similarities between the generated explanation and a reference. 
In our scenario, we want to ensure that the explanations are faithful to the input and that users can
simulate model behavior on new examples based on the explanation. 
We measure faithfulness by using the explanations to make predictions on the inputs, $X_{1:N}$, and evaluating how often the labels suggested by the explanations match with the original predictions from the classifier in question, $y_{1:N}$.\footnote{This approach is also referred to as \textit{fidelity} in the Explainable AI literature \citep{jacovi-goldberg-2020-towards}.}
Similarly, to measure simulatability, we use the explanations to make predictions on unseen examples from the task, $X'_{1:M}$, and assess their alignment with the ground-truth labels, $y'_{1:M}$. 
We use a semantic parser to parse explanations generated for unseen synthetic and real-world tasks.

\begin{table*}[t!]
\normalsize
\begin{center}
\scalebox{0.64}{
\begin{tabular}{ l|l|l} 
 \toprule
 \begin{tabular}{@{}l@{}}\textbf{Task}\\ \textbf{Complexity}\end{tabular} & \begin{tabular}{@{}l@{}}\textbf{Ground-truth}\\ \textbf{Explanations}\end{tabular} & \begin{tabular}{@{}l@{}}\textbf{\model}\\ \textbf{Explanations}\end{tabular} \\
 \midrule
 \multirow{2}{*}{\texttt{simple}}
 & If pdsu lesser than or equal to 1014, then no & If pdsu not greater than 1020, then it is certainly no\\[2pt]
 \cline{2-3} & \rule{0pt}{3ex}If vpgu equal to antartica, then blicket & If vpgu equal to antartica, then it is definitely blicket
\\
 \midrule
 \multirow{3}{*}{\texttt{quantifier}} & If twqk equal to no, then it is seldom fem & If twqk equal to no, then it is seldom fem\\[2pt]\cline{2-3} & \rule{0pt}{3ex}If bgbs not equal to 4, then it is certainly 2 & If bgbs equal to 4, then it is seldom 2\\

 \midrule
\multirow{3}{*}{\texttt{conjunction}} & If aehw equal to no AND hxva equal to africas, then tupa. & If hxva equal to africas, then it is definitely tupa\\ \cline{2-3} & \rule{0pt}{3ex}If kjwx greater than or equal to 18 OR bzjf greater than 1601, then it is definitely 1.  & If kjwx not lesser than 19, then it is likely 1\\ 

\bottomrule
 \end{tabular}
 }
 \end{center}
\vspace{-0.1in}
\caption{ Explanations generated by \model for unseen synthetic tasks for different task complexities.
}
\vspace{-0.1in}
\label{tab:ood-examples}
\end{table*}
\begin{figure*}
    \centering
    \begin{subfigure}[b]{0.32\textwidth}
         \centering
         \includegraphics[width=1.05\textwidth]{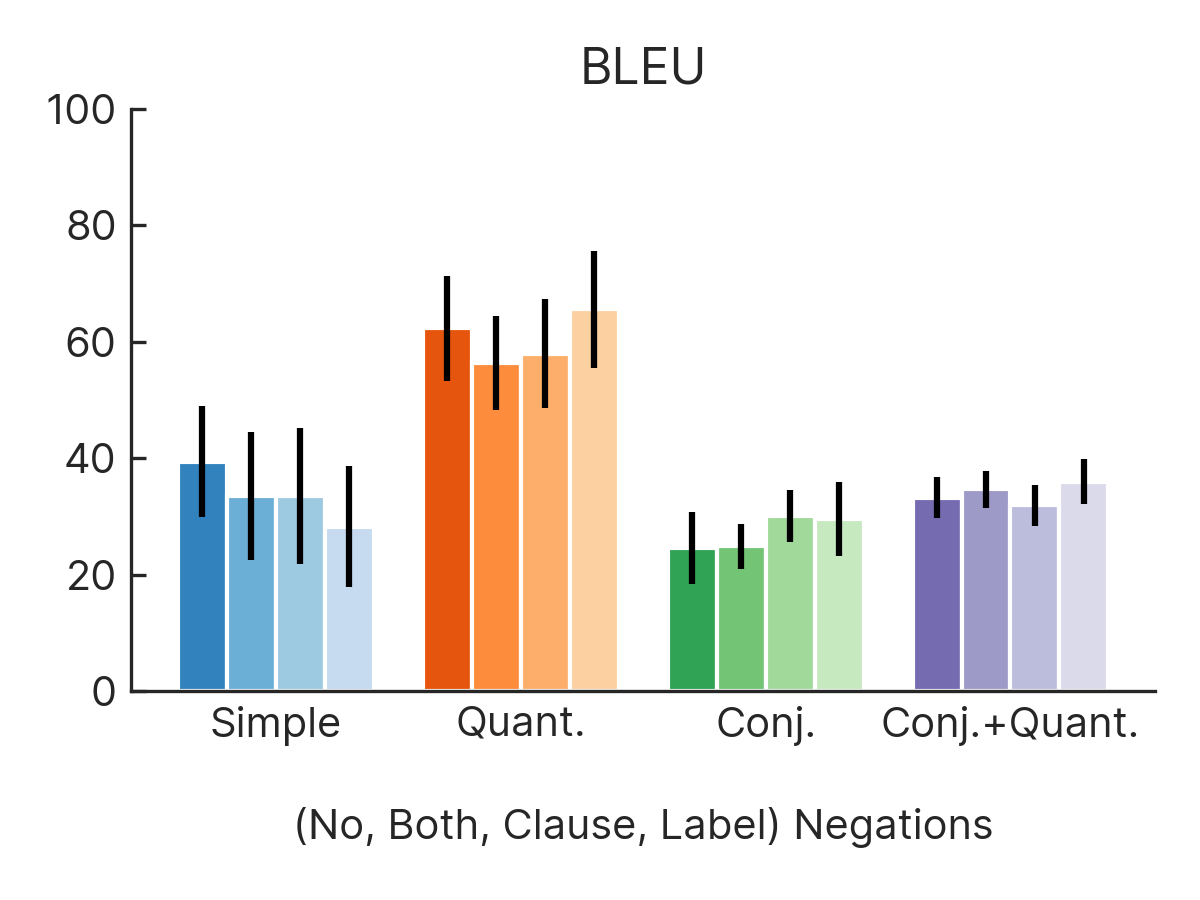}
         \caption{BLEU}
         \label{fig:bleu}
     \end{subfigure}
     \hfill
     \begin{subfigure}[b]{0.32\textwidth}
         \centering
         \includegraphics[width=1.05\textwidth]{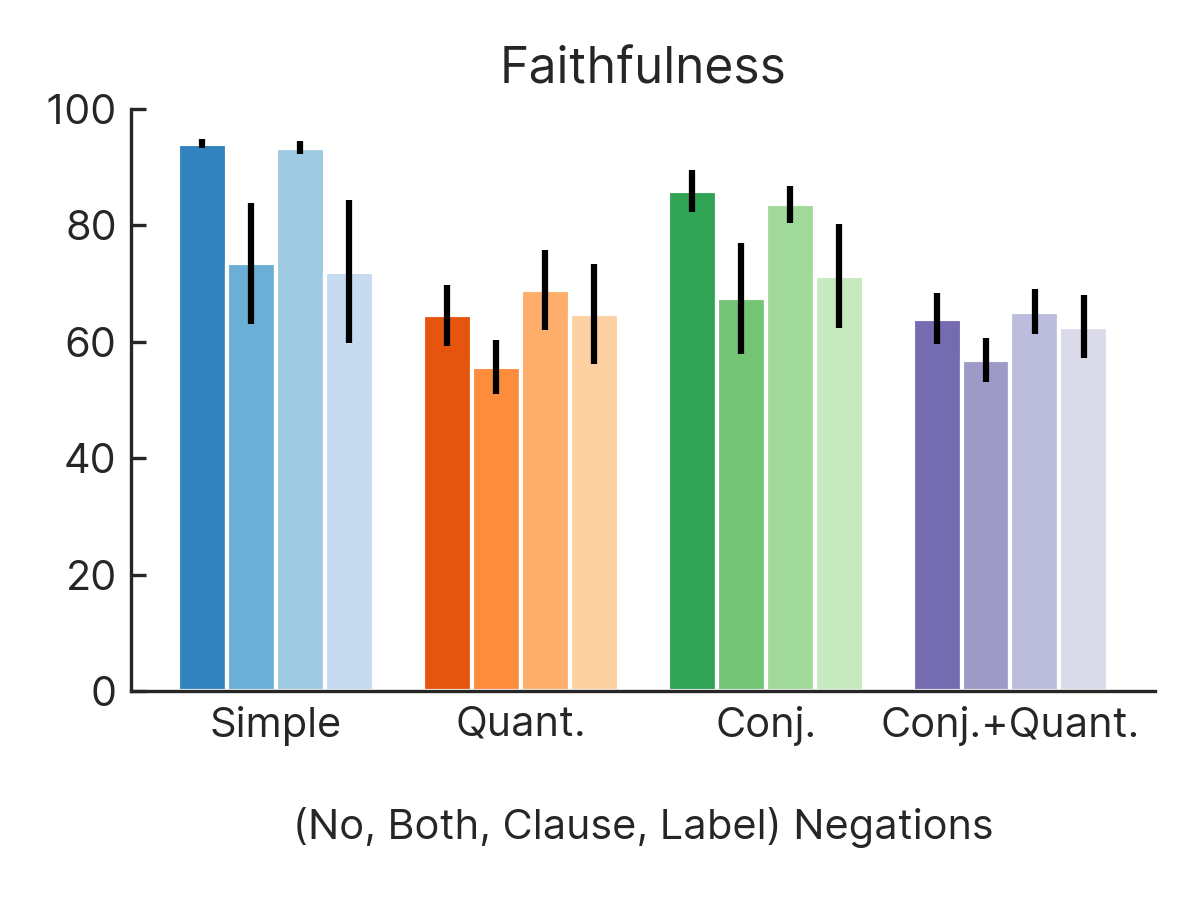}
         \caption{Faithfulness}
         \label{fig:ood_faith}
     \end{subfigure}
     \hfill
     \begin{subfigure}[b]{0.32\textwidth}
         \centering
         \includegraphics[width=1.05\textwidth]{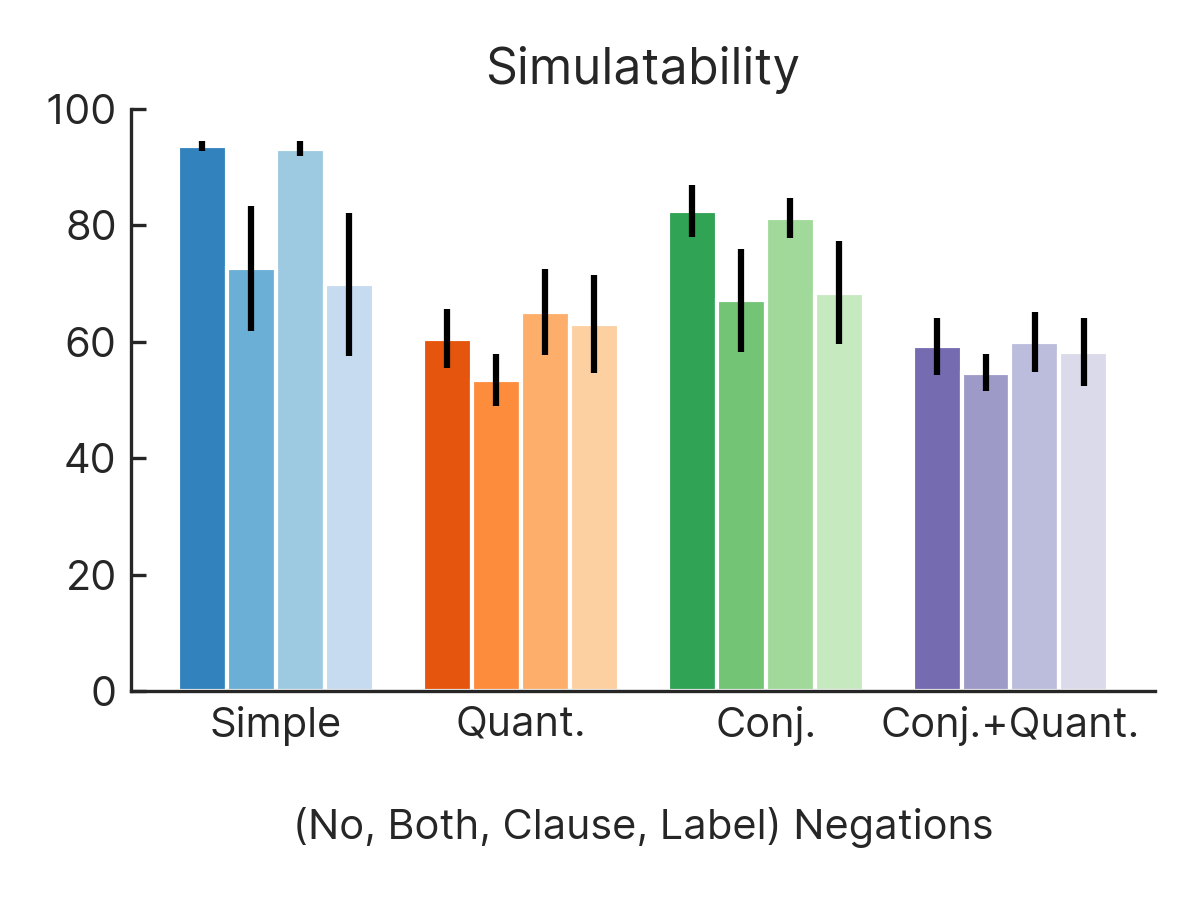}
         \caption{Simulatability}
         \label{fig:ood_sim}
     \end{subfigure}
     \vspace{-0.5em}
    \caption{Results on unseen tasks of different complexities (negations, quantifiers, conjunctions). These are numbers averaged over 20 datasets per task category. Error bars indicate standard deviation. 
    }
    \label{fig:results_ood}
\end{figure*}

\section{Unseen Synthetic Task Results}
\label{sec:results_ood}

Following training, we evaluate \model by generating explanations for 20 synthetic datasets from the different task complexities described in \secref{sec:pretrain_data} (presence/absence of \textit{conjunctions}, \textit{negations}, and \textit{quantifiers}; Table \ref{tab:syn-template}).
These datasets were not used during training and are therefore considered \emph{unseen} tasks. We use greedy decoding to generate the explanations from \model.

Example generations in Table \ref{tab:ood-examples} indicate that while we train \model equally across all complexities, the generations are biased towards using quantifiers in the explanations. 
Consequently, surface-level metrics, such as BLEU, are highest for the generated explanations in the quantifier tasks category (see Figure \ref{fig:bleu}). 
The generated explanations seldom contain conjunctions leading to lower faithfulness and simulatability measures on datasets that contain conjunctive explanations in Figure \ref{fig:results_ood}. 
Further, generated explanations never have negations in the label, i.e., no \texttt{`not \{label\}'} explanations. Hence, the faithfulness and simulatability are lower than the no-negation datasets (Figure~\ref{fig:results_ood}; second and fourth bars in each of the four sets).
\section{Explaining Datasets with \model}
\label{sec:real_eval}

We investigate the efficacy of \model explanations compared to human-written explanations for aiding in classifying examples from datasets in the \cluesreal benchmark \cite{menon-etal-2022-clues}. 
Specifically,
we evaluate the accuracy of \quexent \cite{ghosh2022quexent}, a recent explanation-guided classifier for the benchmark,\footnote{FLAN-T5-XL \cite{chung-etal-2022-flant5}, an effective model at learning from instructions, underperformed \quexent here.} 
using both explanations across three binary classification tasks from the benchmark's zero-shot evaluation set. 
Explanations justifying the labeling rationale for datasets in \cluesreal were developed by individuals that were shown a few examples.
Hence, this benchmark provides an ideal setting for evaluating \model explanations, which are also generated based on patterns observed in a few examples.\footnote{In the implementation, we test the performance of the top-10 generated \model explanations (seeing 10 input examples) with 10 crowd-sourced explanations for each task from \cluesreal.}

\begin{table}[ht!]
\normalsize
\begin{center}
\scalebox{0.8}{
\begin{tabular}{ l|l|r|r} 
 \toprule
 \textbf{Task} & \begin{tabular}{@{}l@{}}\textbf{\model}\\ \textbf{Decoding} \\\textbf{Strategy}\end{tabular} & \textbf{Accuracy} & \begin{tabular}{@{}r@{}}\textbf{Human} \\\textbf{Explanation} \\\textbf{Accuracy}\end{tabular}\\
 \midrule
 \begin{tabular}{@{}l@{}}\texttt{banknote}\\\texttt{authentication}\end{tabular} & \begin{tabular}{@{}l@{}} Greedy \\\Beam  \\\Perfeat \end{tabular} & \begin{tabular}{@{}r@{}}$50.6$\\ $52.0$\\ $\mathbf{52.4}$\end{tabular} & $56.4$\\
 \midrule
 \begin{tabular}{@{}l@{}}\texttt{indian}\\\texttt{liver patient}\end{tabular} & \begin{tabular}{@{}l@{}} Greedy \\\Beam  \\\Perfeat \end{tabular} & \begin{tabular}{@{}r@{}} $44.6$\\ $40.5$\\ $\mathbf{51.4}$ \end{tabular} & $48.6$\\
 \midrule
 \begin{tabular}{@{}l@{}}\texttt{tic-tac-toe}\\\texttt{endgame}\end{tabular} & \begin{tabular}{@{}l@{}} Greedy \\\Beam  \\\Perfeat \end{tabular} & \begin{tabular}{@{}r@{}} $\mathbf{63.0}$\\ $56.8$\\ $57.8$ \end{tabular} & $66.1$\\
\bottomrule
 \end{tabular}
 }
 \end{center}
\vspace{-0.1in}
\caption{ 
Simulatability of \model explanations as measured by \quexent~\cite{ghosh2022quexent}. 
Accuracy measures \quexent's ability to predict ground-truth labels.
}
% \rrm{Need to pick datasets}
\vspace{-0.1in}
\label{tab:clues_quexent}
\end{table}

Results in Table~\ref{tab:clues_quexent} show the the performance of \quexent with the best \model explanations is able to get within $7\%$ of the performance with crowd-sourced explanations across tasks.
Additionally, when optimizing for faithfulness, diversifying the pool of candidate explanations is helpful as indicated by the consistent performance improvement of \Perfeat decoding over beam search.

We provide qualitative examples of generated explanations for different tasks in Table \ref{tab:clues-examples}. 
While \model can recover some crowd-sourced explanations, we also observed hallucinations and a lack of numerical understanding leading to errors in explanations. 
We leave these errors for future studies to investigate and address. 
\section{Interpreting Classifiers with \model}
\label{sec:interpret_classifiers}
In this section, we compare the quality of \model explanations to two popular explanation methods in \lime and \anchors using simulated (\secref{sec:simuser_exp}) and human user studies (\secref{sec:human_exp}). 
Before the experiments, we briefly describe the baselines (\secref{sec:baseline_exp_methods}) and  datasets (\secref{sec:interpret_datasets}) that we use in our experimentation.

\subsection{Baseline Explanation Methods}
\label{sec:baseline_exp_methods}
\paragraph{LIME.}
\citet{ribeiro2016should} approximates model behavior locally by fitting a linear model to the outputs of the black box model on perturbations sampled from a distribution around an input. 
The linear model output is the probability of an example belonging to the same class as the input used for constructing the explanation. 
The quality of \lime explanations depends on the number of perturbations sampled around an input.
Typically, \lime uses $5000$ perturbations to fit a good linear model and generate explanations for tabular datasets.

\paragraph{Anchors.} 
\citet{ribeiro2018anchors} proposed a technique for building rule lists that predict model behavior with high precision, i.e., the label mentioned in the explanation matches the model prediction with a high likelihood when the rules apply. 
To achieve high precision, \anchors often sacrifices explanation coverage, i.e., the number of instances where the rules apply may be very limited. 
\anchors employs a PAC (probably approximately correct) learning approach to generate a rule list based on samples from a distribution around an input. 
Unlike \lime which uses input perturbations, \anchors uses training set samples of the black-box model that is close to the input for its construction.

\subsection{Datasets and Models}
\label{sec:interpret_datasets}
Following previous work \cite{ribeiro2018anchors}, we perform experiments for three structured classification tasks. 
The first is the {\tt adult} dataset from the UCI-ML repository \cite{Dua:2019}, where the objective is to predict the income level of a person (more or less than $\$50,000$).
The second dataset used is the {\tt recidivism} dataset \cite{schmidt1988predicting}, where the task is to predict if a convict is likely to re-commit crimes.
The third dataset is the {\tt travel-insurance} dataset \cite{travel-insurance} obtained from Kaggle, where the task is to predict if a traveler will purchase travel insurance.

In all experiments, we use five features from the available set of features for each dataset.\footnote{For each dataset, we use the top-5 features that maximize information between labels and examples in the training set.}
To ensure consistency, we follow the data-processing scheme and model architectures used in \citet{ribeiro2018anchors}.
For each dataset, we train and explain four classifiers: logistic regression, decision trees, neural networks, and gradient-boosted trees. 

We report the faithfulness and simulatability of explanations generated by different methods. 
Here, simulatability is measured as the proportion of test set examples for which the model prediction matches the label in the explanation.

\subsection{Automated Evaluation} \label{sec:simuser_exp}
In this section, we conduct simulated user studies to compare the effectiveness of \lime, \anchors, and \model in generating explanations for classifiers.
\paragraph{Setup.}
For each classifier-dataset pair, we subsample $100$ random subsets from the validation set, each with $10$ examples and the corresponding predictions from the classifier. 
Next, for each subset, we generate explanations using \lime, \anchors, and the different \model variants. 
For \lime and \anchors, we compute the submodular pick (SP) variant of the explanations \citep{ribeiro2016should}.

As mentioned in \secref{sec:baseline_exp_methods}, \lime and \anchors need to sample examples or perturbations around the input to generate high-quality explanations. 
However, \model generates explanations without any additional information beyond the examples from the subset. 
To perform a fair evaluation, we report results for a budget-constrained setting, wherein methods can make a maximum of 15 classifier calls. 
This corresponds to performing 1 perturbation per example for \lime and using 5 training examples for \anchors. 
Budget-constrained scenarios form a realistic setting in the current landscape where black-box classifiers, like GPT-3, are expensive to query both monetarily and computationally.

\begin{table}[ht!]
\normalsize
\begin{center}
% \subfloat[\texttt{adult}]{
\scalebox{0.61}{
\begin{tabular}{ ll|rr|rr|rr|rr} 
 \toprule
 & & \multicolumn{2}{c|}{\textbf{lr}} & \multicolumn{2}{c|}{\textbf{dt}} & \multicolumn{2}{c|}{\textbf{nn}} & \multicolumn{2}{c}{\textbf{xgb}}\\
 & \textbf{explanation} & 
 \textbf{faith} & \textbf{sim} & \textbf{faith} & \textbf{sim}  & \textbf{faith} & \textbf{sim} &  \textbf{faith} & \textbf{sim} \\
 \midrule

\parbox[t]{2mm}{\multirow{5}{*}{\rotatebox[origin=c]{90}{\texttt{\textbf{adult}}}}} & LIME & $51.3$ & $50.2$ & $53.4$ & $49.8$ & $53.9$ & $50.8$ & $52.8$ & $50.0$\\
& Anchor & $\mathbf{80.2}$ & $\mathbf{70.9}$ & $57.8$ & $52.9$ & $57.3$ & $50.7$ & $55.3$ & $52.1$\\
& \model & $56.3$ & $49.2$ & $55.8$ & $49.2$ & $56.2$ & $49.6$ & $57.1$ & $49.3$\\
& \model-BS & $67.6$ & $52.9$ & $67.1$ & $51.9$ & $67.5$ & $52.6$ & $68.4$ & $51.4$\\
& \model-PF & $74.3$ & $57.4$ & $\mathbf{74.4}$ & $\mathbf{54.6}$ & $\mathbf{75.1}$ & $\mathbf{56.5}$ & $\mathbf{75.0}$ & $\mathbf{55.9}$\\
\midrule
\parbox[t]{2mm}{\multirow{5}{*}{\rotatebox[origin=c]{90}{\texttt{\textbf{travel ins.}}}}} & LIME & $53.0$ & $50.6$ & $54.9$ & $54.0$ & $52.6$ & $51.3$ & $53.0$ & $50.4$\\
& Anchor & $60.6$ & $\mathbf{69.6}$ & $\mathbf{73.5}$ & $\mathbf{71.7}$ & $61.3$ & $\mathbf{69.2}$ & $45.5$ & $51.1$\\
& \model & $58.7$ & $61.1$ & $56.4$ & $63.0$ & $57.5$ & $60.0$ & $55.1$ & $52.8$\\
& \model-BS & $\mathbf{73.3}$ & $63.5$ & $71.7$ & $63.0$ & $\mathbf{72.7}$ & $62.5$ & $68.8$ & $\mathbf{54.1}$\\
& \model-PF & $72.6$ & $60.2$ & $71.7$ & $60.5$ & $72.1$ & $60.3$ & $\mathbf{71.8}$ & $53.4$\\
\midrule
\parbox[t]{2mm}{\multirow{5}{*}{\rotatebox[origin=c]{90}{\texttt{\textbf{recidivism}}}}} & LIME & $53.9$ & $50.0$ & $51.4$ & $50.0$ & $51.7$ & $50.0$ & $50.7$ & $50.0$\\
& Anchor & $58.5$ & $\mathbf{60.6}$ & $\mathbf{76.1}$ & $\mathbf{58.8}$ & $\mathbf{77.4}$ & $\mathbf{58.5}$ & $\mathbf{76.1}$ & $\mathbf{58.9}$\\
& \model & $53.9$ & $50.6$ & $54.3$ & $51.9$ & $52.4$ & $51.4$ & $53.9$ & $51.7$\\
& \model-BS & $69.7$ & $51.3$ & $69.1$ & $52.8$ & $68.8$ & $52.7$ & $69.1$ & $52.5$\\
& \model-PF & $\mathbf{70.9}$ & $51.6$ & $69.3$ & $51.6$ & $69.7$ & $51.3$ & $69.4$ & $51.1$\\
\bottomrule
\end{tabular}
}
% }
\end{center}
\vspace{-0.1in}
\caption{
Faithfulness and simulatability when executing different explanations for three datasets. Results are averaged over 100 runs (or subsets). \textbf{Bold} numbers indicate the best explanation for a particular classifier-metric pair.
BS = beam search, PF = \Perfeat decoding.
}
\vspace{-0.1in}
\label{tab:main-sim-recidivism}
\end{table}

\paragraph{Results.}
In Table \ref{tab:main-sim-recidivism}, we see that \lime falls short of \anchors and \model variants on all classifier-dataset combinations, likely caused by the small number of perturbations available to \lime.
Among different \model decoding strategies, the results suggest that faithfulness improves with better decoding strategies, with \model-PF having the best performance overall.
Overall, we observe that across the three datasets, \model-PF is more faithful than \lime and \anchors by $37\%$ and $11\%$, respectively, highlighting the utility of our approach in this budget-constrained scenario.

To address scenarios where many examples are accessible cheaply, in \secref{sec:more_examples}, we explore ways to incorporate more examples to improve the quality of \model explanations.

\subsection{Human Evaluation}
\label{sec:human_exp}

In user studies, we measure the ability of users to interpret explanations and predict the behavior of models trained on the {\tt adult} dataset.
We use the full budget \lime and \anchors explanations and the \model-\Perfeat explanations from the previous section.
In a pilot study, we found workers had difficulty interpreting the meaning of different quantifiers.
Hence, we post-process \model explanations by converting quantifiers to confidence values based on previous work \cite{menon-etal-2022-clues}.\footnote{Fig.~\ref{fig:expl_show} shows examples of post-processed explanations.}

\begin{figure}
    \centering
    \begin{subfigure}[b]{0.235\textwidth}
         \centering
         \includegraphics[width=\textwidth]{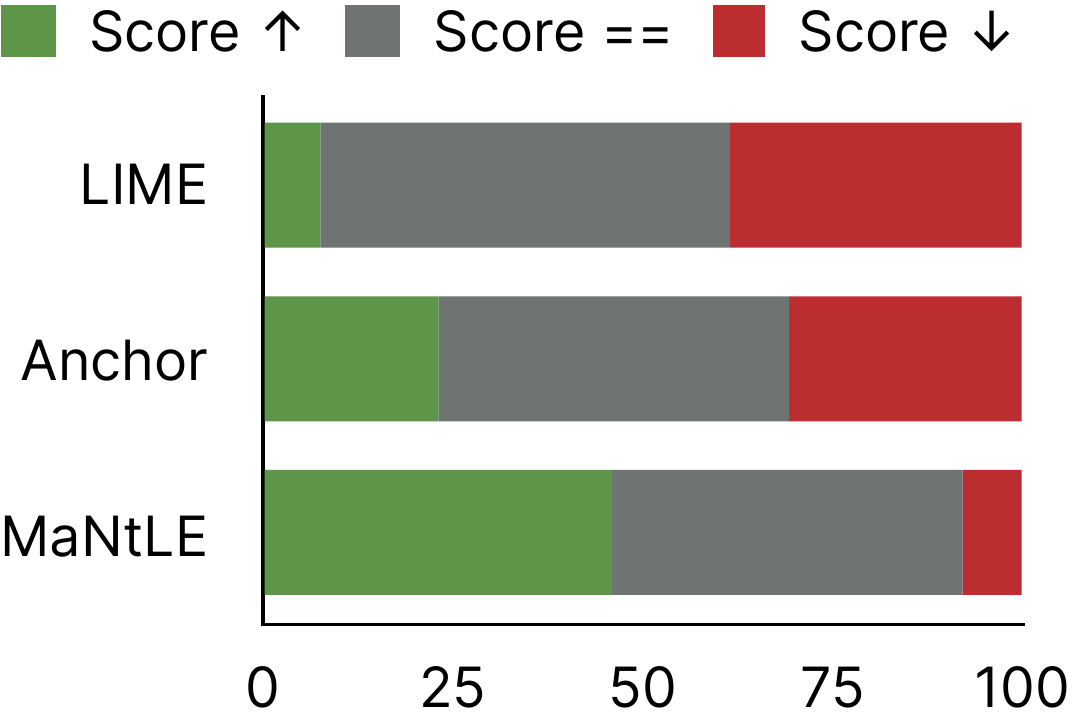}
         \caption{Perturbation}
         \label{fig:he_perturb}
     \end{subfigure}
     \begin{subfigure}[b]{0.235\textwidth}
         \centering
         \includegraphics[width=\textwidth]{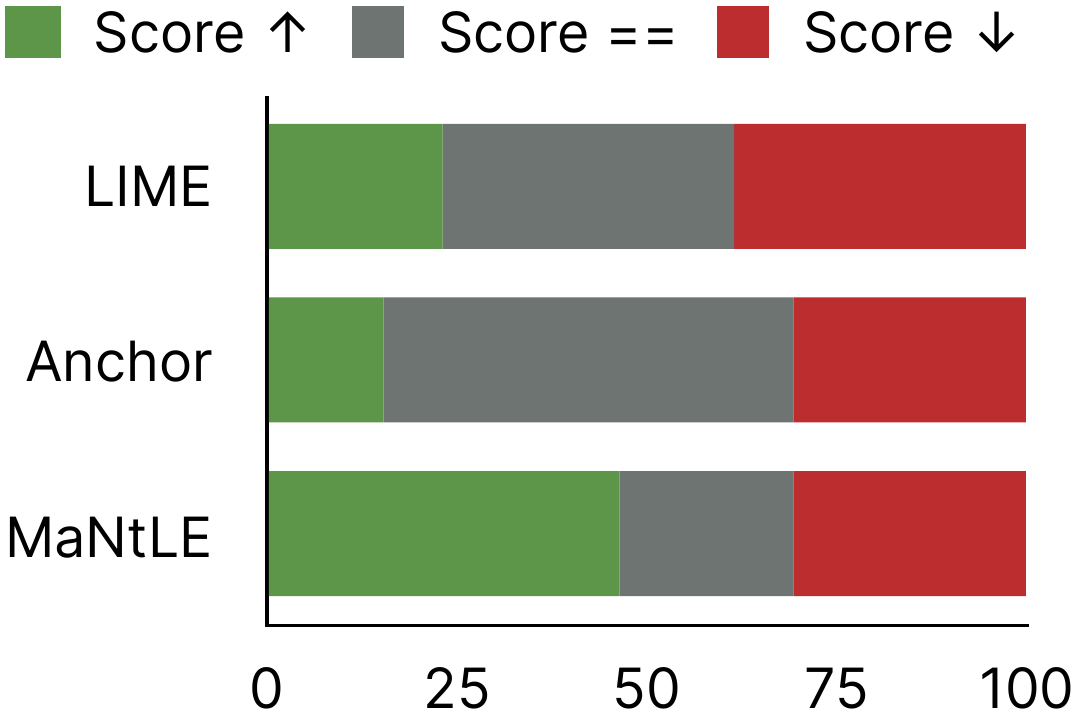}
         \caption{Simulation}
         \label{fig:he_simulation}
     \end{subfigure}
    \caption{Percentage of instances where workers understanding of model behavior improved (\textcolor{OliveGreen}{\textbf{green}}), declined (\textcolor{BrickRed}{\textbf{red}}), and did not change (\textcolor{darkgray}{\textbf{gray}}) on reviewing different explanations for the {\tt adult} dataset.}
    \label{fig:human_eval}
\end{figure}

\paragraph{Which explanations help users predict model behavior?} 
In Figure \ref{fig:human_eval}, we present the results of our study.
We report the percentage of instances where the user predictions of model behavior improve, worsen, or remain the same on perturbed examples (perturbation) and test examples (simulation) after reviewing explanations, following the setup in \citet{hase-bansal-2020-evaluating}. 
23 workers took part in this study conducted on Amazon Mechanical Turk and were compensated at $\$12$/hr.

In our results (Figure \ref{fig:human_eval}), we found that users can predict model behavior more accurately in 46\% of cases after reviewing \model explanations, compared to 15\% for \lime and 19\% for \anchors. 
Additionally, users were less likely to make more incorrect predictions of model behavior after viewing \model explanations, with only 19\% of cases, compared to 38\% for \lime and 31\% for \anchors. 
Hence, our explanations are more reliable in helping users predict model behavior.

\paragraph{Which explanations would general practitioners prefer?}
For this study, we recruited 25 participants on Prolific
% \footnote{\url{https://www.prolific.co/}}
who are currently pursuing at least an undergraduate degree to rate explanations on a 1-5 Likert scale for understandability, informativeness, and overall preference.\footnote{Fig.~\ref{fig:temp_sub} shows how we define the scale for each property.}
We chose this demographic to reflect the expected diversity of industry experts' educational backgrounds likely to use explanations to better understand their systems. 

\begin{figure}
    \centering
    \includegraphics[width=0.48\textwidth]{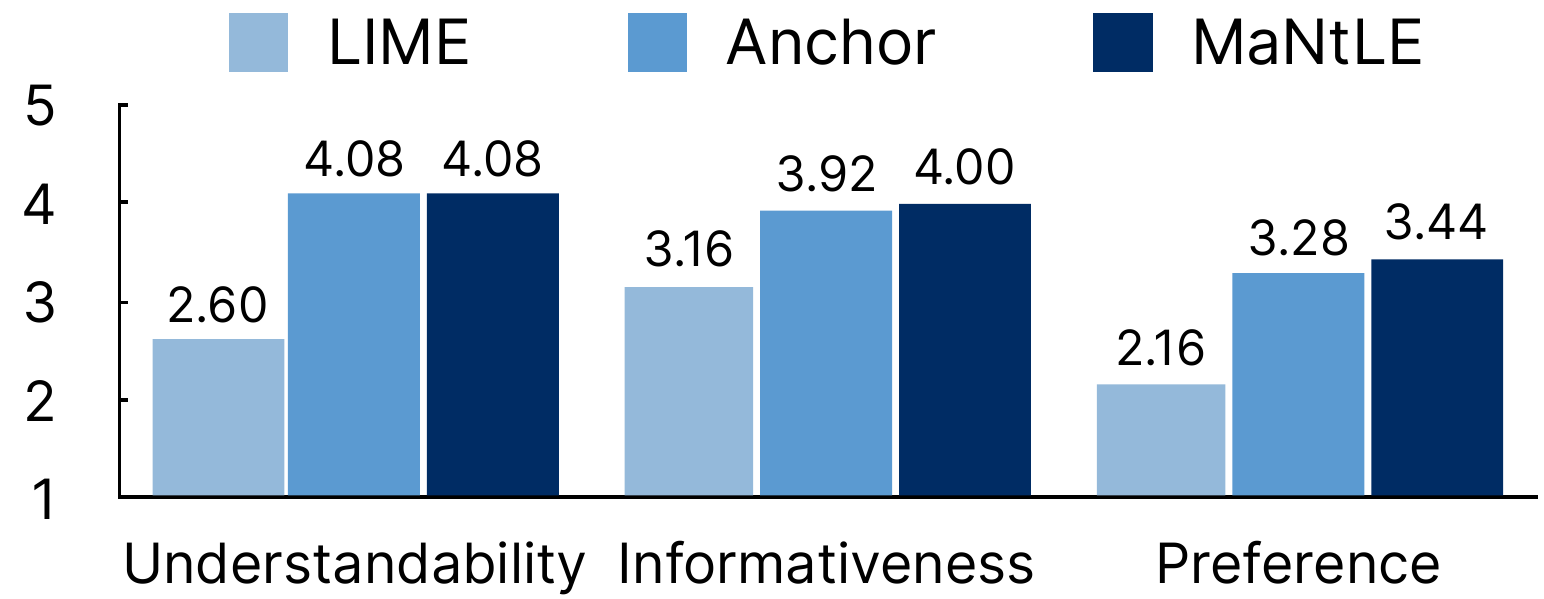}
    \caption{Preference for different explanation techniques by workers with at least undergraduate level education, as indicated by Likert ratings on a 1-5 scale.}
    \label{fig:likert}
\end{figure}

Results in Fig.~\ref{fig:likert} indicate that workers struggled to comprehend attribution scores from \lime compared to \model. 
A paired-sample t-test for overall preference revealed significance ($2.16$~vs.~$3.44$; $p$-value~$<0.001$). 
In contrast, workers found \anchors informative, but their low coverage hindered their overall preference compared to \model. 

Notably, workers prefer \model explanations for their clarity and simplicity, citing phrases such as \textit{`clearly defines information'} and \textit{`uses more layman terms'}.
Nevertheless, some expressed concerns that the explanations were \textit{`not descriptive enough'}, \textit{`\dots semi-specific when compared to \lime'}. 
This suggests that explanations clarifying how each feature affects the classifier's decision-making process could improve user understanding and satisfaction.

\section{Analysis}
In this section, we analyze \model based on two key aspects: (a) factors affecting the strong generalization of \model, and (b) stability of \model when the number of input examples is varied.

\subsection{How does scale affect the performance of \model?}
\label{sec:model_data_ablate}
\begin{figure}[!ht]
    \centering
    \includegraphics[width=0.49\textwidth]{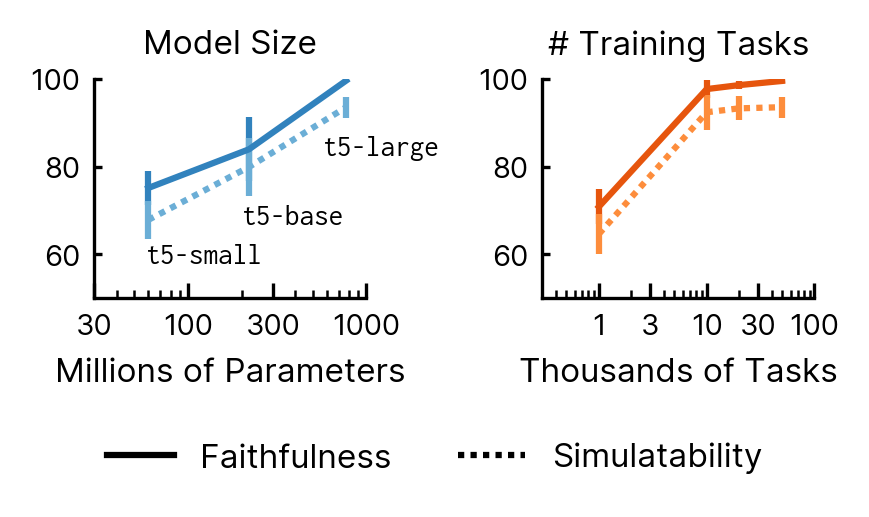}
    \vspace{-0.5cm}
    \caption{Faithfulness and simulatability performance of \model on held-out tasks with increase in the scale of model size (left) and dataset size (right). Error bars indicate standard deviation.
    }
    \label{fig:model_ablation}
\end{figure}

Here, we evaluate \model's performance by scaling model parameters and training tasks. 
To accomplish this, we create a synthetic benchmark consisting of 50K datasets, each with ground-truth explanations using conjunctions, which are challenging for \model to learn. 
For evaluation, 20 datasets from the benchmark were held-out and we measure the faithfulness and simulatability of explanations generated by fine-tuned models on these datasets. 
We fine-tune a \texttt{T5-Large} model using nearly all 50K datasets for model scale experiments, and vary tasks between 1K to 50K for task scale experiments.

To study the effect of model scale, we fine-tune different variants of \texttt{T5}, ranging from \texttt{T5-Small} to \texttt{T5-Large}.
In Figure \ref{fig:model_ablation} (left), we see that increasing the scale of models from \texttt{T5-Small} to \texttt{T5-Large} significantly improves the faithfulness and simulatability of generated explanations.
Further, increasing the number of training tasks improves both metrics, as shown in Figure \ref{fig:model_ablation} (right). 
Notably, fine-tuning a \texttt{T5-Large} model on smaller number of tasks ($1$K) leads to poorer performance than a \texttt{T5-Small} model fine-tuned on larger number of tasks ($50$K), even when trained with the same hyperparameters for the same duration. 
Taken together, expanding datasets and increasing model sizes further could improve \model's performance.

\subsection{Can \model take advantage of more examples?}
\label{sec:more_examples}
The maximum number of tokens allowed in the \texttt{T5} encoder limits the input capacity of \model.
This restricts \model when a large number of examples are available for the classifier being explained. 

To address this challenge, we propose a technique that enables \model-Beam and \model-PF to handle more input examples. 
The method involves dividing a set of $N$ examples into eight subsets of $10$ examples each, using them to generate explanations, and selecting the explanation with the highest ``simulatability'' score among all $N$ examples as the best explanation.
Finally, we report the faithfulness and simulatability of our explanations using this approach to explain logistic regression and decision tree classifiers for the {\tt adult} dataset.
\begin{figure}[t!]
    \centering
    \includegraphics[width=0.49\textwidth]{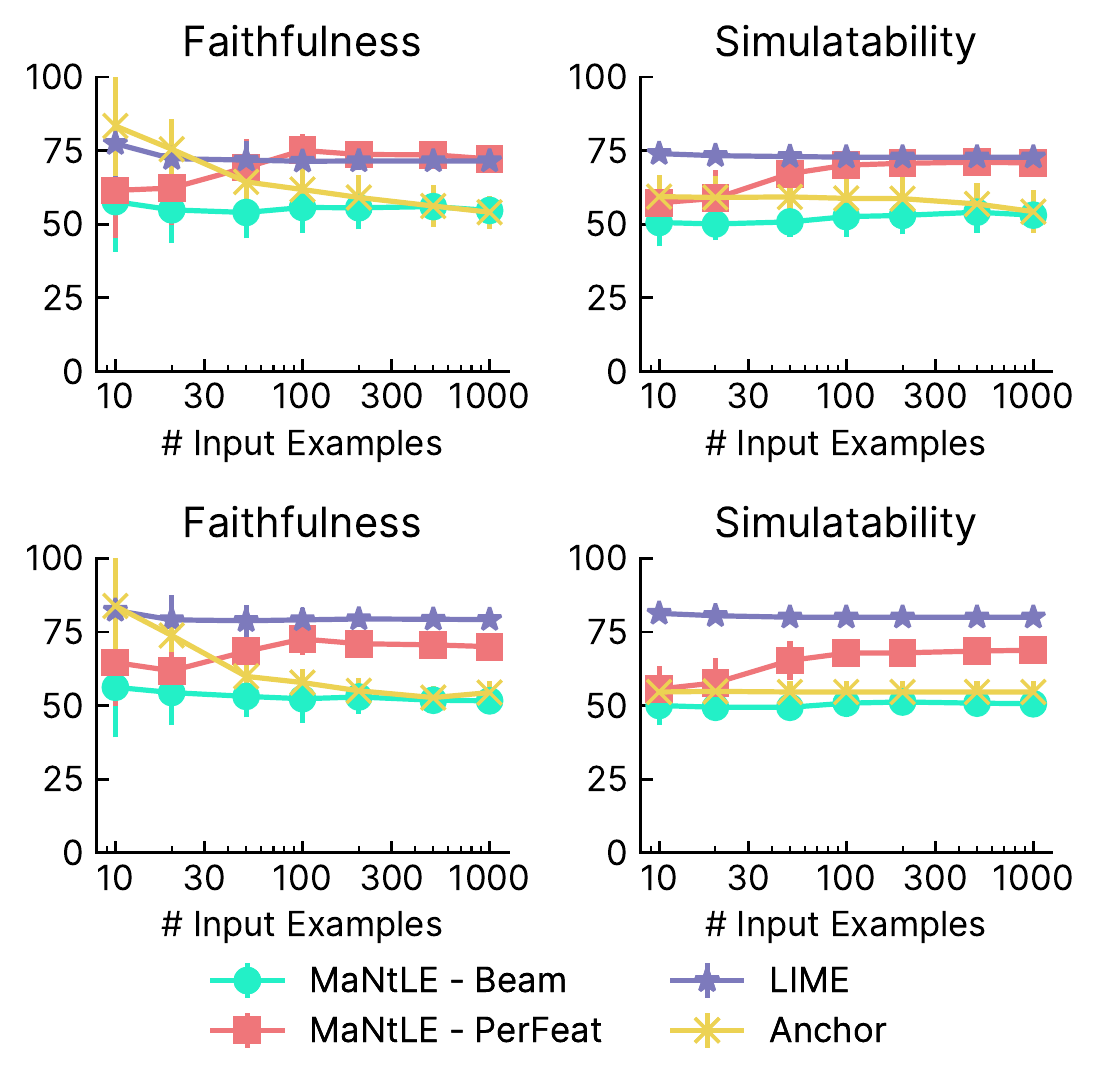}
    \vspace{-1em}
    \caption{\textbf{Increasing number of input examples improves simulatability and faithfulness of \model-PF explanations.} While the simulatability of LIME, Anchor, and \model remain constant with an increase in the number of input examples, explanations from \model-PF improve simulatability and faithfulness. Here, we compute metrics over 100 runs for two classifiers (\textit{logistic regression -- top row}, \textit{decision tree -- bottom row}) trained on the {\tt adult} dataset. 
    \lime and \anchors shown here are full-budget submodular pick variants. Error bars show standard deviation.}
    \label{fig:sampleexp_main}
\end{figure}

In Figure \ref{fig:sampleexp_main}, we see that the proposed procedure improves the faithfulness and simulatability measures of \model-PF.
We also find that \model-PF performance is comparable the full-budget submodular pick variant of \lime, indicating that our approach allows \model to match \lime in explanation quality when provided access to a large number of examples.
The procedure, however, does not improve the metrics for \model-Beam which implies that the diversity of generated explanations is essential.
Incorporating more than eight subsets could further improve the performance of \model-PF. We leave this for future work to explore.

\section{Conclusion}

In this work, we introduce \model, a model-agnostic natural language explainer, that generates faithful explanations for structured classifiers. 
We use recent insights in massive multi-task training to train models that can generate classifier rationale. 
As \model can explain classifiers simply by inspecting predictions made by classifiers, it can be used in a model-agnostic fashion similar to popular explanation methods like \lime and \anchors. 
In simulation studies and human evaluations, we show that \model explanations are more faithful than \lime and comparable in faithfulness to \anchors on multiple datasets. 
Our work suggests the potential for natural language interfaces to enhance end-user trust and promote the safe usage of ML models in critical tasks by providing explanations of the classifier decision-making process.
Future work can look to extend our work to develop ``patches'' \cite{murty-etal-2022-fixing} for improving classifiers, refining decoding techniques for more faithful explanations, and integrating more complex reasoning patterns in generated explanations.
\section{Limitations}

Our method is exclusively designed to explain classifiers operating on structured datasets. Utilizing \model for other input types, such as raw text and images, is out of the scope of this work. 

Moreover, as we have pointed out in our experiments, being a neural text generation model, \model suffers from hallucinations and lacks numerical understanding, as a result of which some generated explanations may be incorrect.

Further, the number of examples that can be packed into the encoder of the \model is limited to 1024 tokens (limit of \texttt{T5} encoder). While our work mentions additional strategies to circumvent this issue, future work could look into additional methods for packing more examples into the input to improve the faithfulness and simulatability of generated explanations. 

Additionally, the kind of logic that can be represented by the outputs of \model is likely limited to the ones seen during training. Hence, we may never observe explanations with nested conjunctions. Future work can identify solutions to incorporate more complex reasoning in explanations. Integrating such reasoning without training \model from scratch is also an interesting future direction.

\bibliography{anthology, custom}

\begin{thebibliography}{35}
\expandafter\ifx\csname natexlab\endcsname\relax\def\natexlab#1{#1}\fi

\bibitem[{Brown et~al.(2020)Brown, Mann, Ryder, Subbiah, Kaplan, Dhariwal,
  Neelakantan, Shyam, Sastry, Askell, Agarwal, Herbert-Voss, Krueger, Henighan,
  Child, Ramesh, Ziegler, Wu, Winter, Hesse, Chen, Sigler, Litwin, Gray, Chess,
  Clark, Berner, McCandlish, Radford, Sutskever, and
  Amodei}]{NEURIPS2020_1457c0d6}
Tom Brown, Benjamin Mann, Nick Ryder, Melanie Subbiah, Jared~D Kaplan, Prafulla
  Dhariwal, Arvind Neelakantan, Pranav Shyam, Girish Sastry, Amanda Askell,
  Sandhini Agarwal, Ariel Herbert-Voss, Gretchen Krueger, Tom Henighan, Rewon
  Child, Aditya Ramesh, Daniel Ziegler, Jeffrey Wu, Clemens Winter, Chris
  Hesse, Mark Chen, Eric Sigler, Mateusz Litwin, Scott Gray, Benjamin Chess,
  Jack Clark, Christopher Berner, Sam McCandlish, Alec Radford, Ilya Sutskever,
  and Dario Amodei. 2020.
\newblock \href
  {https://proceedings.neurips.cc/paper/2020/file/1457c0d6bfcb4967418bfb8ac142f64a-Paper.pdf}
  {Language models are few-shot learners}.
\newblock In \emph{Advances in Neural Information Processing Systems},
  volume~33, pages 1877--1901. Curran Associates, Inc.

\bibitem[{Chopra et~al.(2019)Chopra, Tessler, and Goodman}]{Chopra2019TheFC}
Sahil Chopra, Michael~Henry Tessler, and Noah~D. Goodman. 2019.
\newblock \href {https://cogsci.mindmodeling.org/2019/papers/0060/index.html}
  {The first crank of the cultural ratchet: Learning and transmitting concepts
  through language}.
\newblock In \emph{CogSci}.

\bibitem[{Chung et~al.(2022)Chung, Hou, Longpre, Zoph, Tay, Fedus, Li, Wang,
  Dehghani, Brahma, Webson, Gu, Dai, Suzgun, Chen, Chowdhery, Narang, Mishra,
  Yu, Zhao, Huang, Dai, Yu, Petrov, Chi, Dean, Devlin, Roberts, Zhou, Le, and
  Wei}]{chung-etal-2022-flant5}
Hyung~Won Chung, Le~Hou, Shayne Longpre, Barret Zoph, Yi~Tay, William Fedus,
  Eric Li, Xuezhi Wang, Mostafa Dehghani, Siddhartha Brahma, Albert Webson,
  Shixiang~Shane Gu, Zhuyun Dai, Mirac Suzgun, Xinyun Chen, Aakanksha
  Chowdhery, Sharan Narang, Gaurav Mishra, Adams Yu, Vincent Zhao, Yanping
  Huang, Andrew Dai, Hongkun Yu, Slav Petrov, Ed~H. Chi, Jeff Dean, Jacob
  Devlin, Adam Roberts, Denny Zhou, Quoc~V. Le, and Jason Wei. 2022.
\newblock \href {https://arxiv.org/abs/2210.11416} {Scaling
  instruction-finetuned language models}.

\bibitem[{Dada et~al.(2019)Dada, Bassi, Chiroma, Abdulhamid, Adetunmbi, and
  Ajibuwa}]{DADA2019e01802}
Emmanuel~Gbenga Dada, Joseph~Stephen Bassi, Haruna Chiroma, Shafi'i~Muhammad
  Abdulhamid, Adebayo~Olusola Adetunmbi, and Opeyemi~Emmanuel Ajibuwa. 2019.
\newblock \href {https://doi.org/https://doi.org/10.1016/j.heliyon.2019.e01802}
  {Machine learning for email spam filtering: review, approaches and open
  research problems}.
\newblock \emph{Heliyon}, 5(6):e01802.

\bibitem[{Dua and Graff(2017)}]{Dua:2019}
Dheeru Dua and Casey Graff. 2017.
\newblock \href {http://archive.ics.uci.edu/ml} {{UCI} machine learning
  repository}.

\bibitem[{Ghosh et~al.(2022)Ghosh, Menon, and Srivastava}]{ghosh2022quexent}
Sayan Ghosh, Rakesh~R Menon, and Shashank Srivastava. 2022.
\newblock Lasque: Improved zero-shot classification from explanations through
  quantifier modeling and curriculum learning.
\newblock \emph{ArXiv}, abs/2212.09104.

\bibitem[{Hase and Bansal(2020)}]{hase-bansal-2020-evaluating}
Peter Hase and Mohit Bansal. 2020.
\newblock \href {https://doi.org/10.18653/v1/2020.acl-main.491} {Evaluating
  explainable {AI}: Which algorithmic explanations help users predict model
  behavior?}
\newblock In \emph{Proceedings of the 58th Annual Meeting of the Association
  for Computational Linguistics}, pages 5540--5552, Online. Association for
  Computational Linguistics.

\bibitem[{Honovich et~al.(2022)Honovich, Shaham, Bowman, and
  Levy}]{honovich2022instruction}
Or~Honovich, Uri Shaham, Samuel~R Bowman, and Omer Levy. 2022.
\newblock Instruction induction: From few examples to natural language task
  descriptions.
\newblock \emph{arXiv preprint arXiv:2205.10782}.

\bibitem[{Jacovi and Goldberg(2020)}]{jacovi-goldberg-2020-towards}
Alon Jacovi and Yoav Goldberg. 2020.
\newblock \href {https://doi.org/10.18653/v1/2020.acl-main.386} {Towards
  faithfully interpretable {NLP} systems: How should we define and evaluate
  faithfulness?}
\newblock In \emph{Proceedings of the 58th Annual Meeting of the Association
  for Computational Linguistics}, pages 4198--4205, Online. Association for
  Computational Linguistics.

\bibitem[{Kim et~al.(2018)Kim, Wattenberg, Gilmer, Cai, Wexler, Viegas, and
  sayres}]{pmlr-v80-kim18d}
Been Kim, Martin Wattenberg, Justin Gilmer, Carrie Cai, James Wexler, Fernanda
  Viegas, and Rory sayres. 2018.
\newblock \href {https://proceedings.mlr.press/v80/kim18d.html}
  {Interpretability beyond feature attribution: Quantitative testing with
  concept activation vectors ({TCAV})}.
\newblock In \emph{Proceedings of the 35th International Conference on Machine
  Learning}, volume~80 of \emph{Proceedings of Machine Learning Research},
  pages 2668--2677. PMLR.

\bibitem[{Lakkaraju et~al.(2022)Lakkaraju, Slack, Chen, Tan, and
  Singh}]{lakkaraju2022rethinking}
Himabindu Lakkaraju, Dylan Slack, Yuxin Chen, Chenhao Tan, and Sameer Singh.
  2022.
\newblock Rethinking {E}xplainability as a {D}ialogue: {A} {P}ractitioner's
  {P}erspective.
\newblock \emph{arXiv preprint arXiv:2202.01875}.

\bibitem[{Lin(2004)}]{lin-2004-rouge}
Chin-Yew Lin. 2004.
\newblock \href {https://aclanthology.org/W04-1013} {{ROUGE}: A package for
  automatic evaluation of summaries}.
\newblock In \emph{Text Summarization Branches Out}, pages 74--81, Barcelona,
  Spain. Association for Computational Linguistics.

\bibitem[{Loshchilov and Hutter(2019)}]{loshchilov2018decoupled}
Ilya Loshchilov and Frank Hutter. 2019.
\newblock \href {https://openreview.net/forum?id=Bkg6RiCqY7} {Decoupled weight
  decay regularization}.
\newblock In \emph{International Conference on Learning Representations}.

\bibitem[{Lundberg and Lee(2017)}]{lundberg2017unified}
Scott~M Lundberg and Su-In Lee. 2017.
\newblock \href
  {https://proceedings.neurips.cc/paper/2017/file/8a20a8621978632d76c43dfd28b67767-Paper.pdf}
  {A {U}nified {A}pproach to {I}nterpreting {M}odel {P}redictions}.
\newblock In \emph{Advances in Neural Information Processing Systems},
  volume~30. Curran Associates, Inc.

\bibitem[{Menon et~al.(2022)Menon, Ghosh, and
  Srivastava}]{menon-etal-2022-clues}
Rakesh~R. Menon, Sayan Ghosh, and Shashank Srivastava. 2022.
\newblock \href {https://doi.org/10.18653/v1/2022.acl-long.451} {{CLUES}: A
  benchmark for learning classifiers using natural language explanations}.
\newblock In \emph{Proceedings of the 60th Annual Meeting of the Association
  for Computational Linguistics (Volume 1: Long Papers)}, pages 6523--6546,
  Dublin, Ireland. Association for Computational Linguistics.

\bibitem[{Mishra et~al.(2022)Mishra, Khashabi, Baral, and
  Hajishirzi}]{mishra-etal-2022-cross}
Swaroop Mishra, Daniel Khashabi, Chitta Baral, and Hannaneh Hajishirzi. 2022.
\newblock \href {https://doi.org/10.18653/v1/2022.acl-long.244} {Cross-task
  generalization via natural language crowdsourcing instructions}.
\newblock In \emph{Proceedings of the 60th Annual Meeting of the Association
  for Computational Linguistics (Volume 1: Long Papers)}, pages 3470--3487,
  Dublin, Ireland. Association for Computational Linguistics.

\bibitem[{Murty et~al.(2022)Murty, Manning, Lundberg, and
  Ribeiro}]{murty-etal-2022-fixing}
Shikhar Murty, Christopher Manning, Scott Lundberg, and Marco~Tulio Ribeiro.
  2022.
\newblock \href {https://aclanthology.org/2022.emnlp-main.797} {Fixing model
  bugs with natural language patches}.
\newblock In \emph{Proceedings of the 2022 Conference on Empirical Methods in
  Natural Language Processing}, pages 11600--11613, Abu Dhabi, United Arab
  Emirates. Association for Computational Linguistics.

\bibitem[{Narang et~al.(2020)Narang, Raffel, Lee, Roberts, Fiedel, and
  Malkan}]{narang2020wt5}
Sharan Narang, Colin Raffel, Katherine Lee, Adam Roberts, Noah Fiedel, and
  Karishma Malkan. 2020.
\newblock Wt5?! training text-to-text models to explain their predictions.
\newblock \emph{arXiv preprint arXiv:2004.14546}.

\bibitem[{Papineni et~al.(2002)Papineni, Roukos, Ward, and
  Zhu}]{10.3115/1073083.1073135}
Kishore Papineni, Salim Roukos, Todd Ward, and Wei-Jing Zhu. 2002.
\newblock \href {https://doi.org/10.3115/1073083.1073135} {Bleu: A method for
  automatic evaluation of machine translation}.
\newblock In \emph{Proceedings of the 40th Annual Meeting on Association for
  Computational Linguistics}, ACL '02, page 311–318, USA. Association for
  Computational Linguistics.

\bibitem[{Paszke et~al.(2019)Paszke, Gross, Massa, Lerer, Bradbury, Chanan,
  Killeen, Lin, Gimelshein, Antiga, Desmaison, Kopf, Yang, DeVito, Raison,
  Tejani, Chilamkurthy, Steiner, Fang, Bai, and Chintala}]{NEURIPS2019_9015}
Adam Paszke, Sam Gross, Francisco Massa, Adam Lerer, James Bradbury, Gregory
  Chanan, Trevor Killeen, Zeming Lin, Natalia Gimelshein, Luca Antiga, Alban
  Desmaison, Andreas Kopf, Edward Yang, Zachary DeVito, Martin Raison, Alykhan
  Tejani, Sasank Chilamkurthy, Benoit Steiner, Lu~Fang, Junjie Bai, and Soumith
  Chintala. 2019.
\newblock \href
  {http://papers.neurips.cc/paper/9015-pytorch-an-imperative-style-high-performance-deep-learning-library.pdf}
  {Pytorch: An imperative style, high-performance deep learning library}.
\newblock In H.~Wallach, H.~Larochelle, A.~Beygelzimer, F.~d\textquotesingle
  Alch\'{e}-Buc, E.~Fox, and R.~Garnett, editors, \emph{Advances in Neural
  Information Processing Systems 32}, pages 8024--8035. Curran Associates, Inc.

\bibitem[{Raffel et~al.(2020)Raffel, Shazeer, Roberts, Lee, Narang, Matena,
  Zhou, Li, Liu et~al.}]{raffel2020exploring}
Colin Raffel, Noam Shazeer, Adam Roberts, Katherine Lee, Sharan Narang, Michael
  Matena, Yanqi Zhou, Wei Li, Peter~J Liu, et~al. 2020.
\newblock Exploring the limits of transfer learning with a unified text-to-text
  transformer.
\newblock \emph{J. Mach. Learn. Res.}, 21(140):1--67.

\bibitem[{Rajani et~al.(2019)Rajani, McCann, Xiong, and
  Socher}]{rajani-etal-2019-explain}
Nazneen~Fatema Rajani, Bryan McCann, Caiming Xiong, and Richard Socher. 2019.
\newblock \href {https://doi.org/10.18653/v1/P19-1487} {Explain yourself!
  leveraging language models for commonsense reasoning}.
\newblock In \emph{Proceedings of the 57th Annual Meeting of the Association
  for Computational Linguistics}, pages 4932--4942, Florence, Italy.
  Association for Computational Linguistics.

\bibitem[{Ribeiro et~al.(2016)Ribeiro, Singh, and Guestrin}]{ribeiro2016should}
Marco~Tulio Ribeiro, Sameer Singh, and Carlos Guestrin. 2016.
\newblock "{W}hy {S}hould {I} {T}rust {Y}ou?" {E}xplaining the {P}redictions of
  {A}ny {C}lassifier.
\newblock In \emph{Proceedings of the 22nd ACM SIGKDD international conference
  on knowledge discovery and data mining}, pages 1135--1144.

\bibitem[{Ribeiro et~al.(2018)Ribeiro, Singh, and
  Guestrin}]{ribeiro2018anchors}
Marco~Tulio Ribeiro, Sameer Singh, and Carlos Guestrin. 2018.
\newblock Anchors: High-precision model-agnostic explanations.
\newblock In \emph{Proceedings of the AAAI conference on artificial
  intelligence}, volume~32.

\bibitem[{Sanh et~al.(2022)Sanh, Webson, Raffel, Bach, Sutawika, Alyafeai,
  Chaffin, Stiegler, Raja, Dey, Bari, Xu, Thakker, Sharma, Szczechla, Kim,
  Chhablani, Nayak, Datta, Chang, Jiang, Wang, Manica, Shen, Yong, Pandey,
  Bawden, Wang, Neeraj, Rozen, Sharma, Santilli, Fevry, Fries, Teehan, Scao,
  Biderman, Gao, Wolf, and Rush}]{sanh2022multitask}
Victor Sanh, Albert Webson, Colin Raffel, Stephen Bach, Lintang Sutawika, Zaid
  Alyafeai, Antoine Chaffin, Arnaud Stiegler, Arun Raja, Manan Dey, M~Saiful
  Bari, Canwen Xu, Urmish Thakker, Shanya~Sharma Sharma, Eliza Szczechla,
  Taewoon Kim, Gunjan Chhablani, Nihal Nayak, Debajyoti Datta, Jonathan Chang,
  Mike Tian-Jian Jiang, Han Wang, Matteo Manica, Sheng Shen, Zheng~Xin Yong,
  Harshit Pandey, Rachel Bawden, Thomas Wang, Trishala Neeraj, Jos Rozen,
  Abheesht Sharma, Andrea Santilli, Thibault Fevry, Jason~Alan Fries, Ryan
  Teehan, Teven~Le Scao, Stella Biderman, Leo Gao, Thomas Wolf, and Alexander~M
  Rush. 2022.
\newblock \href {https://openreview.net/forum?id=9Vrb9D0WI4} {Multitask
  prompted training enables zero-shot task generalization}.
\newblock In \emph{International Conference on Learning Representations}.

\bibitem[{Scao et~al.(2022)Scao, Fan, Akiki, Pavlick, Ili{\'c}, Hesslow,
  Castagn{\'e}, Luccioni, Yvon, Gall{\'e} et~al.}]{scao2022bloom}
Teven~Le Scao, Angela Fan, Christopher Akiki, Ellie Pavlick, Suzana Ili{\'c},
  Daniel Hesslow, Roman Castagn{\'e}, Alexandra~Sasha Luccioni, Fran{\c{c}}ois
  Yvon, Matthias Gall{\'e}, et~al. 2022.
\newblock Bloom: A 176b-parameter open-access multilingual language model.
\newblock \emph{arXiv preprint arXiv:2211.05100}.

\bibitem[{Schmidt and Witte(1988)}]{schmidt1988predicting}
Peter Schmidt and Ann~D Witte. 1988.
\newblock \emph{Predicting recidivism in north carolina, 1978 and 1980}.
\newblock Inter-university Consortium for Political and Social Research.

\bibitem[{Shi et~al.(2022)Shi, Tse, Luo, D'Addona, and Pau}]{credit-risk}
Si~Shi, Rita Tse, Wuman Luo, Stefano D'Addona, and Giovanni Pau. 2022.
\newblock \href {https://doi.org/10.1007/s00521-022-07472-2} {Machine
  learning-driven credit risk: a systemic review}.
\newblock \emph{Neural Computing and Applications}, 34(17):14327--14339.

\bibitem[{Singh et~al.(2022)Singh, Morris, Aneja, Rush, and
  Gao}]{singh2022explaining}
Chandan Singh, John~X Morris, Jyoti Aneja, Alexander~M Rush, and Jianfeng Gao.
  2022.
\newblock Explaining patterns in data with language models via interpretable
  autoprompting.
\newblock \emph{arXiv preprint arXiv:2210.01848}.

\bibitem[{Srivastava et~al.(2018)Srivastava, Labutov, and
  Mitchell}]{srivastava-etal-2018-zero}
Shashank Srivastava, Igor Labutov, and Tom Mitchell. 2018.
\newblock \href {https://doi.org/10.18653/v1/P18-1029} {Zero-shot learning of
  classifiers from natural language quantification}.
\newblock In \emph{Proceedings of the 56th Annual Meeting of the Association
  for Computational Linguistics (Volume 1: Long Papers)}, pages 306--316,
  Melbourne, Australia. Association for Computational Linguistics.

\bibitem[{Sundararajan et~al.(2017)Sundararajan, Taly, and
  Yan}]{pmlr-v70-sundararajan17a}
Mukund Sundararajan, Ankur Taly, and Qiqi Yan. 2017.
\newblock \href {https://proceedings.mlr.press/v70/sundararajan17a.html}
  {Axiomatic attribution for deep networks}.
\newblock In \emph{Proceedings of the 34th International Conference on Machine
  Learning}, volume~70 of \emph{Proceedings of Machine Learning Research},
  pages 3319--3328. PMLR.

\bibitem[{Tejashvi(2019)}]{travel-insurance}
Tejashvi. 2019.
\newblock \href
  {https://www.kaggle.com/datasets/tejashvi14/travel-insurance-prediction-data}
  {Travel insurance prediction data}.

\bibitem[{Wei et~al.(2022)Wei, Bosma, Zhao, Guu, Yu, Lester, Du, Dai, and
  Le}]{wei2022finetuned}
Jason Wei, Maarten Bosma, Vincent Zhao, Kelvin Guu, Adams~Wei Yu, Brian Lester,
  Nan Du, Andrew~M. Dai, and Quoc~V Le. 2022.
\newblock \href {https://openreview.net/forum?id=gEZrGCozdqR} {Finetuned
  language models are zero-shot learners}.
\newblock In \emph{International Conference on Learning Representations}.

\bibitem[{Wolf et~al.(2020)Wolf, Debut, Sanh, Chaumond, Delangue, Moi, Cistac,
  Rault, Louf, Funtowicz, Davison, Shleifer, von Platen, Ma, Jernite, Plu, Xu,
  Le~Scao, Gugger, Drame, Lhoest, and Rush}]{wolf-etal-2020-transformers}
Thomas Wolf, Lysandre Debut, Victor Sanh, Julien Chaumond, Clement Delangue,
  Anthony Moi, Pierric Cistac, Tim Rault, Remi Louf, Morgan Funtowicz, Joe
  Davison, Sam Shleifer, Patrick von Platen, Clara Ma, Yacine Jernite, Julien
  Plu, Canwen Xu, Teven Le~Scao, Sylvain Gugger, Mariama Drame, Quentin Lhoest,
  and Alexander Rush. 2020.
\newblock \href {https://doi.org/10.18653/v1/2020.emnlp-demos.6} {Transformers:
  State-of-the-art natural language processing}.
\newblock In \emph{Proceedings of the 2020 Conference on Empirical Methods in
  Natural Language Processing: System Demonstrations}, pages 38--45, Online.
  Association for Computational Linguistics.

\bibitem[{Zhang* et~al.(2020)Zhang*, Kishore*, Wu*, Weinberger, and
  Artzi}]{Zhang*2020BERTScore:}
Tianyi Zhang*, Varsha Kishore*, Felix Wu*, Kilian~Q. Weinberger, and Yoav
  Artzi. 2020.
\newblock \href {https://openreview.net/forum?id=SkeHuCVFDr} {Bertscore:
  Evaluating text generation with bert}.
\newblock In \emph{International Conference on Learning Representations}.

\end{thebibliography}
\bibliographystyle{acl_natbib}

\appendix
\section{Experiment Details}

Here we provide implementation details such as hyperparameters, hardware and software used for developing \model and running our experiments.

\subsection{Libraries}
We use the HuggingFace library \cite{wolf-etal-2020-transformers} for all the transformer-based models. For \texttt{T5} models, we experiment with \texttt{t5-small}, \texttt{t5-base}, and \texttt{t5-large}\footnote{\url{https://huggingface.co/t5-large}} across the main experiments and analyses. Our main experiment results are based on the \texttt{t5-large} model. Pre-trained checkpoints for these models are publicly available on the HuggingFace library. All models are coded in PyTorch 1.13.1 \cite{NEURIPS2019_9015}.

\subsection{Pre-training Hyperparameters}
We used the \texttt{t5-large} model and trained it using the AdamW \cite{loshchilov2018decoupled} optimizer with a learning rate of $1e-5$ and weight decay of $1e-2$ for $2,000,000$ steps using the standard language modeling objective. Each gradient step is computed over a batch of 4 samples with no gradient accumulation steps. The maximum length of the input is clipped to 1024 tokens, which roughly corresponds to 10-12 input-prediction pairs being encoded in each sample, while we limit the decoder to generating 64 tokens since that was sufficient to generate the longest explanations from our training set. We chose the best checkpoint based on the performance over generation metrics as well as faithfulness and simulatability on 20 held-out validation tasks.

The model was fine-tuned using full precision on a single NVIDIA A100-PCIE-40GB GPU, 400GB RAM, and 40 CPU cores for $\sim25$ days.

\subsection{Pre-training Datasets} \label{sec:pretrain_data}
Pre-training was performed using programmatically generated datasets whose explanations followed the if-then structure following \citet{menon-etal-2022-clues}. We also utilize the different complexities described in this prior work, which enabled our model to perform diverse types of reasoning. Overall, there were 24 different complexities that varied by: (a) the presence of quantifiers in explanations, (b) the presence of conjunctions in explanations, and (c) the presence of negations in explanations. The quantifiers we adopt in this work, along with their values, follow from prior work in \citet{srivastava-etal-2018-zero}. For conjunctions, we can have tasks with explanations that have nested conjunctions (AND-OR / OR-AND explanations) or simple conjunctions (AND / OR). For negations, there are more subdivisions based on the positioning of the negation. For example, if we have an explanation of the form, \textit{`If a equal to 1 then yes'}, then a clause negation corresponds to an explanation of the form \textit{`If a not equal to 1, then yes'}, and a label negation corresponds to \textit{`If a equal to 1, then not yes'}. Hence, the presence of negations can vary by no negations, only clause negations, only label negations, and negations in clause+label. Overall, we have $2$ (quantifier) $\times$ $3$ (conjunctions) $\times$ $4$ (negations) $= 24$ different complexities. We use $\sim 8,000$ tasks per complexity for training, which leads to the massive pre-training dataset of $\sim200,000$ tasks. Table \ref{tab:syn-template} concisely lists the template used for different task complexities.

\begin{table*}[ht!]
\normalsize
\begin{center}
\scalebox{0.66}{
\begin{tabular}{l|l|l} 
 \toprule
 \textbf{Task Complexity} & \textbf{Template} & \textbf{Example Explanations} \\
 \midrule
 \texttt{simple}
 & If \texttt{\{cond\}}, then \texttt{\{label\}} & If pdsu lesser than or equal to 1014, then no
\\
\texttt{conjunction}
 & If \texttt{\{cond1\}} \textbf{AND/OR} \texttt{\{cond1\}}, then \texttt{\{label\}} & If aehw equal to no AND hxva equal to africas, then tupa
\\
\texttt{clause negation}
 & If \texttt{\{feat\_name\}} \textbf{not} equal to \texttt{\{feat\_value\}}, then \texttt{\{label\}} & If bgbs not equal to 4, then 2 
 \\
 \texttt{label negation}
 & If \texttt{\{cond(s)\}}, then \textbf{not} \texttt{\{label\}} & If aehw equal to yes, then not tupa.
 \\
 \texttt{clause+label negation}
 & If \texttt{\{feat\_name\}} \textbf{not} equal to \texttt{\{feat\_value\}}, then \textbf{not} \texttt{\{label\}} & If szoj not equal to 3, then not 5
 \\
 \texttt{quantifier}
 & If \texttt{\{cond\}}, then it is \textbf{\texttt{\{quantifier\}}} \texttt{\{label\}} & If twqk equal to no, then it is seldom fem
\\
\bottomrule
 \end{tabular}
}
 \end{center}
\vspace{-0.1in}
\caption{Templates and example explanations for different task complexities in the synthetic training benchmark. For brevity, we omit mentions of combinations of complexities, e.g., \texttt{conjunction + quantifier}.}
\vspace{-0.1in}
\label{tab:syn-template}
\end{table*}

\subsection{Real-world Task Explanation Generation}

Owing to the pre-training procedure of \model, which involved feature names as single-word lowercase strings, it is essential that input to \model for real-world tasks is formatted in a similar fashion. Therefore, we transform all feature names from real-world tasks into a format that can be processed by \model, accomplished by eliminating spaces and converting all characters in the feature names to lowercase. To ensure that the generated explanations are understandable to humans, we perform a post-processing step to convert these lowercase feature names back into their original format. This process is crucial for achieving accurate explanations using \model.

\begin{table*}[ht!]
\normalsize
\begin{center}
\scalebox{0.62}{
\begin{tabular}{l|l|l} 
 \toprule
 \textbf{Task} & \textbf{Explanations from \cluesreal} & \textbf{\model-PF Explanations} \\
 \midrule
 \multirow{3}{*}{\texttt{banknote-authentication}}
 & Below 3.80 skewness leads to the original. & If skewness lesser than or equal to 3.049, then it is occasionally Fake.\\[2pt]
 \cline{2-3} & \rule{0pt}{3ex}\multirow{2}{*}{Kurtosis is high value so it is original.} & If kurtosis lesser than or equal to 0.995, then it is often Fake \\ & & \textcolor{red}{If kurtosis lesser than 9600, then it is frequently Fake $\times$}
\\
 \midrule
\multirow{2}{*}{\texttt{indian-liver-patient}} &  The SGPT Low percentage so the liver patient was no & If SGPT lesser than or equal to 39, then patient is generally No\\[2pt]\cline{2-3} & Age group above 40 ensures liver patient & If age lesser than or equal to 39, then patient is generally No\\
\midrule
\multirow{1}{*}{\texttt{tic-tac-toe-endgame}} & \rule{0pt}{4ex}\begin{tabular}{@{}l@{}}Without b categories in middle middle square comes\\ under the Positive group.\end{tabular} & If middle-middle-square equal to x, then Game over is sometimes positive\\
\bottomrule
 \end{tabular}
}
 \end{center}
\vspace{-0.1in}
\caption{Explanations generated by \model for \cluesreal datasets.}
\vspace{-0.1in}
\label{tab:clues-examples}
\end{table*}

\subsection{Model and Dataset Scale Analysis} 

For the model-scale analysis, we fine-tune from a \texttt{t5-large} checkpoint using the conjunction datasets. These models were trained using similar hyperparameters as that from pre-training. However, since the datasets only come from a single complexity, namely conjunctions, these models were trained for $200,000$ steps.

\section{Extended Related Work}
\paragraph{Instruction Generation by Large Language Models.} 
While LLMs can perform many tasks when prompted with instructions or a few examples, their underlying reasoning remains opaque to users.
Some recent works \cite{honovich2022instruction, singh2022explaining} explore techniques to prompt LLMs to generate instructions based on a few examples from synthetic and real-world datasets. 
While our training procedure learns to generate explanations for datasets akin to these prior works, our primary objective is to explain classifiers to understand their classification rationale rather than datasets.

\begin{figure*}[t!]
    \centering
    \begin{subfigure}[b]{0.32\textwidth}
         \centering
         \includegraphics[width=\textwidth]{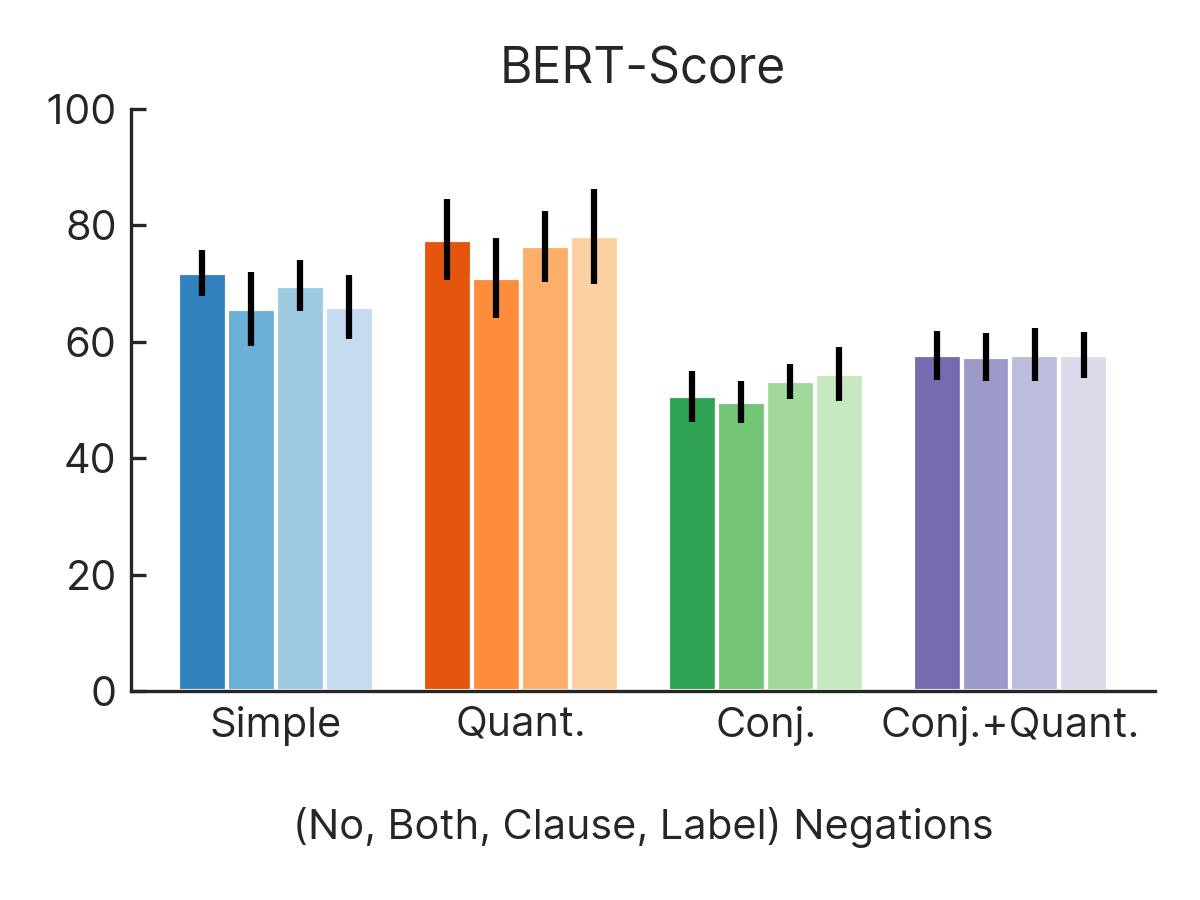}
         \caption{BERT-Score}
         \label{fig:app_bertscore}
     \end{subfigure}
     \hfill
     \begin{subfigure}[b]{0.32\textwidth}
         \centering
         \includegraphics[width=\textwidth]{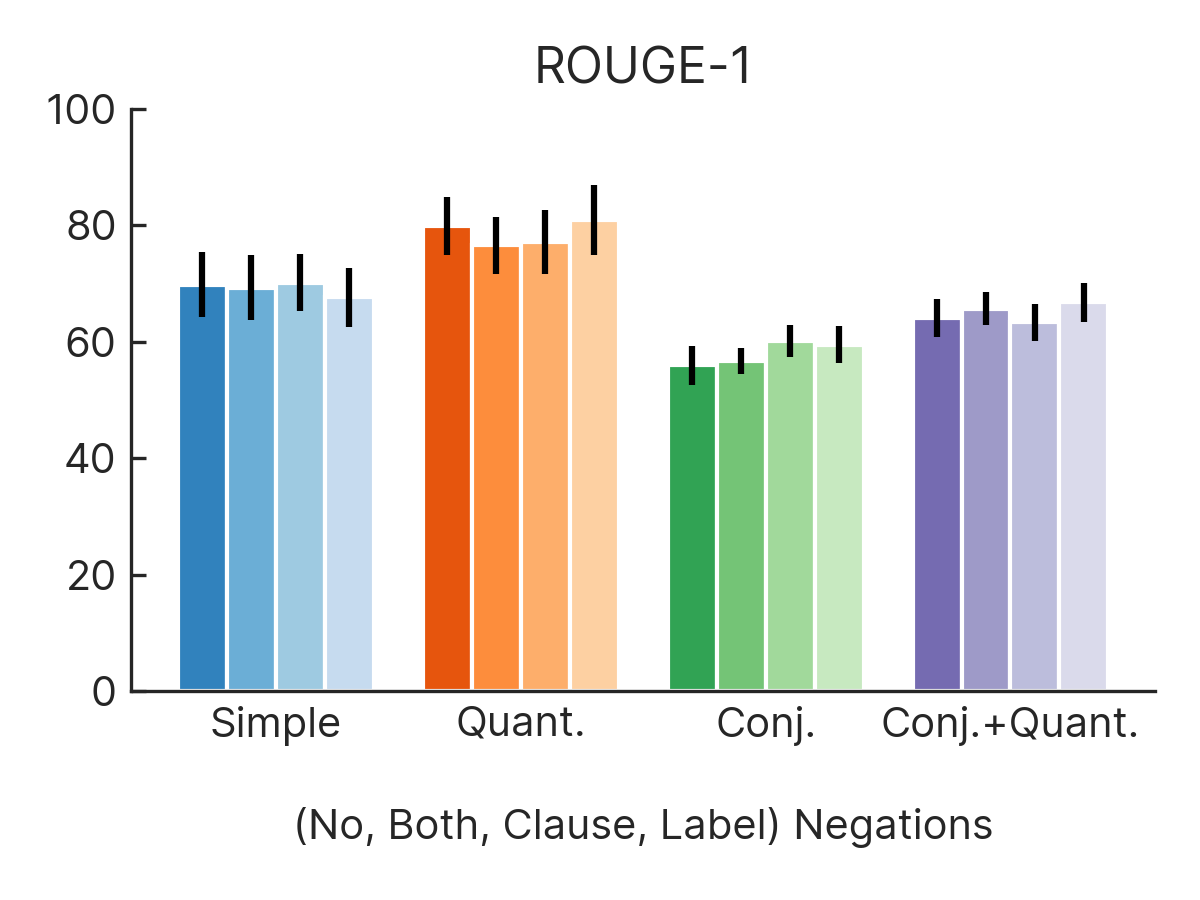}
         \caption{ROUGE-1}
         \label{fig:app_rouge1}
     \end{subfigure}
     \hfill
     \begin{subfigure}[b]{0.32\textwidth}
         \centering
         \includegraphics[width=\textwidth]{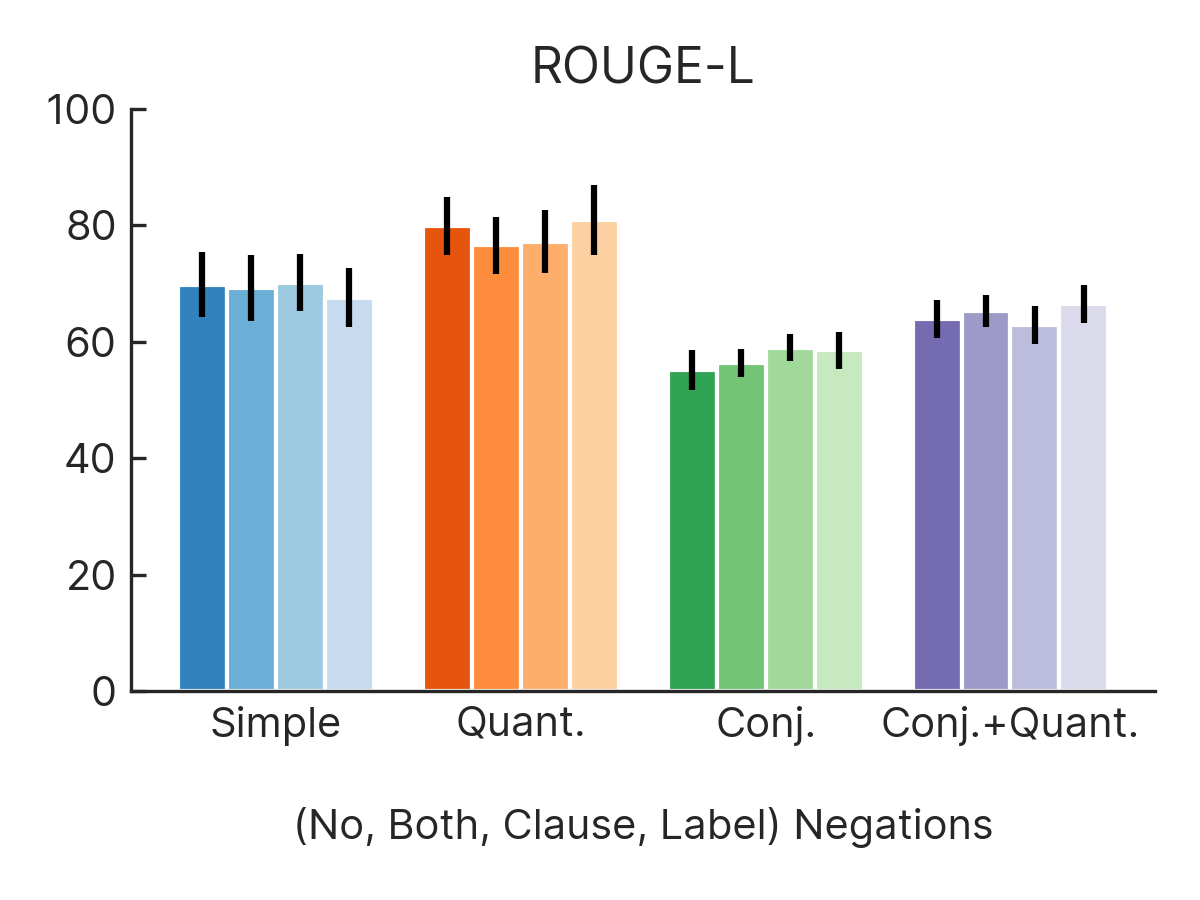}
         \caption{ROUGE-L}
         \label{fig:app_rougel}
     \end{subfigure}
     \begin{subfigure}[b]{0.32\textwidth}
         \centering
         \includegraphics[width=\textwidth]{images/ood_syn/bleu_score.png}
         \caption{BLEU}
         \label{fig:app_bleu}
     \end{subfigure}
     \hfill
     \begin{subfigure}[b]{0.32\textwidth}
         \centering
         \includegraphics[width=\textwidth]{images/ood_syn/seen_faithfulness.png}
         \caption{Faithfulness}
         \label{fig:app_faith}
     \end{subfigure}
     \hfill
     \begin{subfigure}[b]{0.32\textwidth}
         \centering
         \includegraphics[width=\textwidth]{images/ood_syn/faithfulness.png}
         \caption{Simulatability}
         \label{fig:app_sim}
     \end{subfigure}
     \caption{Results on OOD tasks of different complexities (negations, quantifiers, conjunctions). These are numbers averaged over 20 datasets per task category. BERT-Score, ROUGE, and BLEU scores are the highest for Quantifier datasets since our model is more adept at generating content that contains single attribute explanations with quantifiers and negations.}
     \label{fig:app_ood}
\end{figure*}
\section{OOD Synthetic Task Results}

In \secref{sec:results_ood}, we evaluated \model explanations on a set of 20 unseen tasks from the synthetic benchmark. The results of this evaluation, presented in Figure \ref{fig:app_ood}, include \model's performance on the full range of generation metrics, as well as measures of faithfulness and simulatability. As noted previously in \secref{sec:results_ood}, the performance metrics for generation, such as BERT-Score, ROUGE-*, and BLEU, have revealed that \model explanations exhibit a closer alignment to the ground-truth explanations when the latter includes quantifiers. This phenomenon can be attributed to a bias that \model acquires towards generating quantifiers towards the end of the training process, as observed in Table 1. However, our analysis of faithfulness and simulatability scores revealed that \model explanations were most effective on the simplest datasets that lacked complexities such as negations, quantifiers, or conjunctions, in line with our expectations.

\section{Extended Analysis}

\subsection{Can \model take advantage of more examples?}

\begin{figure*}
    \centering
    \begin{subfigure}[b]{0.98\textwidth}
         \centering
         \includegraphics[width=\textwidth]{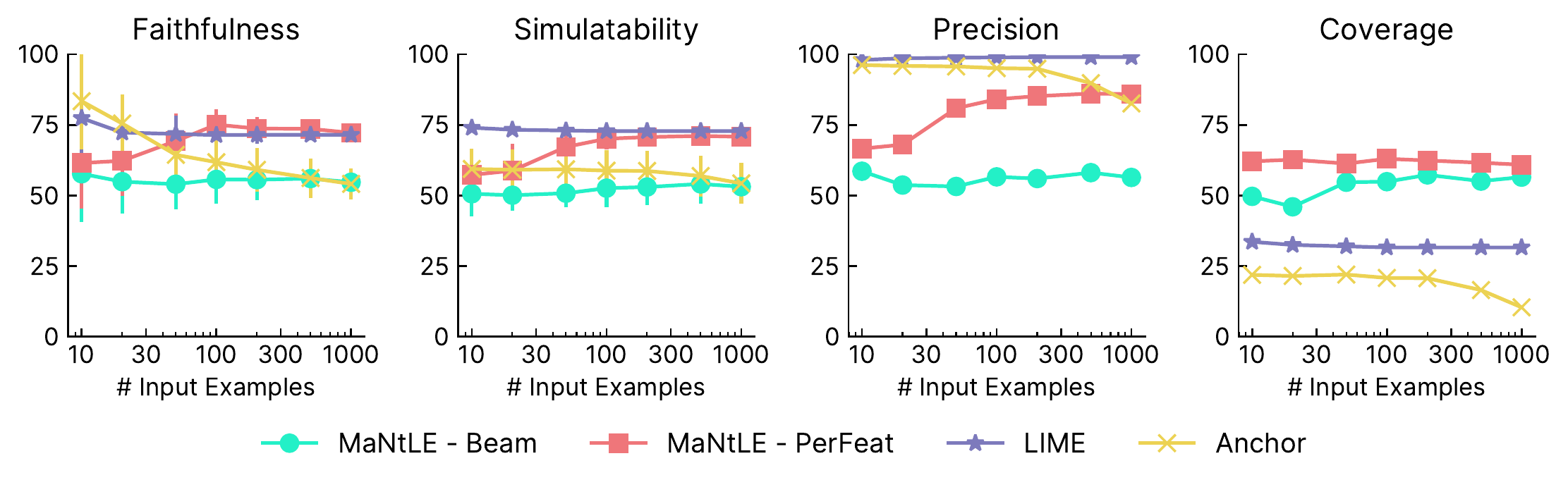}
         \caption{Logistic Regression}
         \label{fig:lr_sampleexp}
     \end{subfigure}
     \begin{subfigure}[b]{0.98\textwidth}
         \centering
         \includegraphics[width=\textwidth]{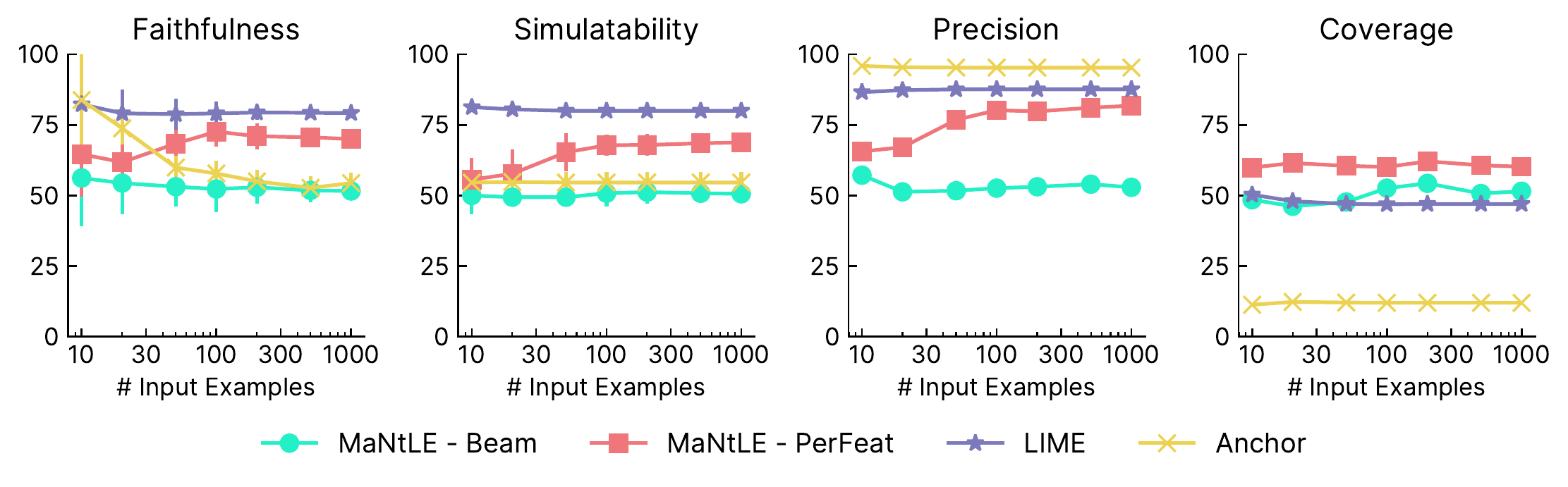}
         \caption{Decision Tree}
         \label{fig:dt_sampleexp}
     \end{subfigure}
     \begin{subfigure}[b]{0.98\textwidth}
         \centering
         \includegraphics[width=\textwidth]{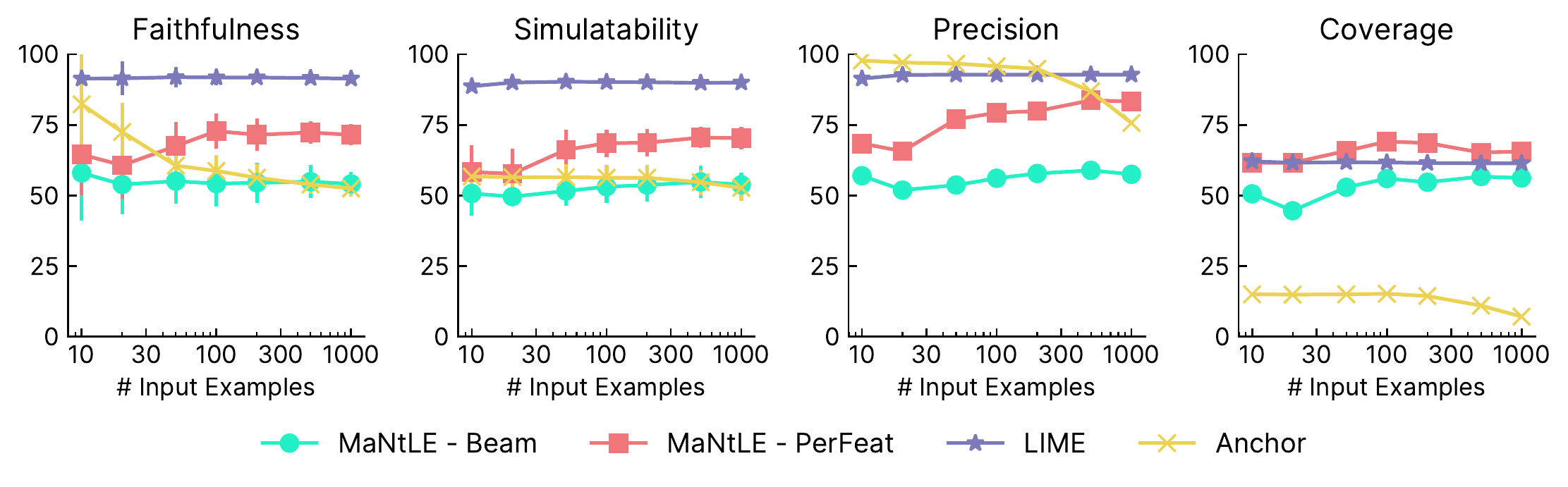}
         \caption{Neural Network}
         \label{fig:nn_sampleexp}
     \end{subfigure}
     \begin{subfigure}[b]{0.98\textwidth}
         \centering
         \includegraphics[width=\textwidth]{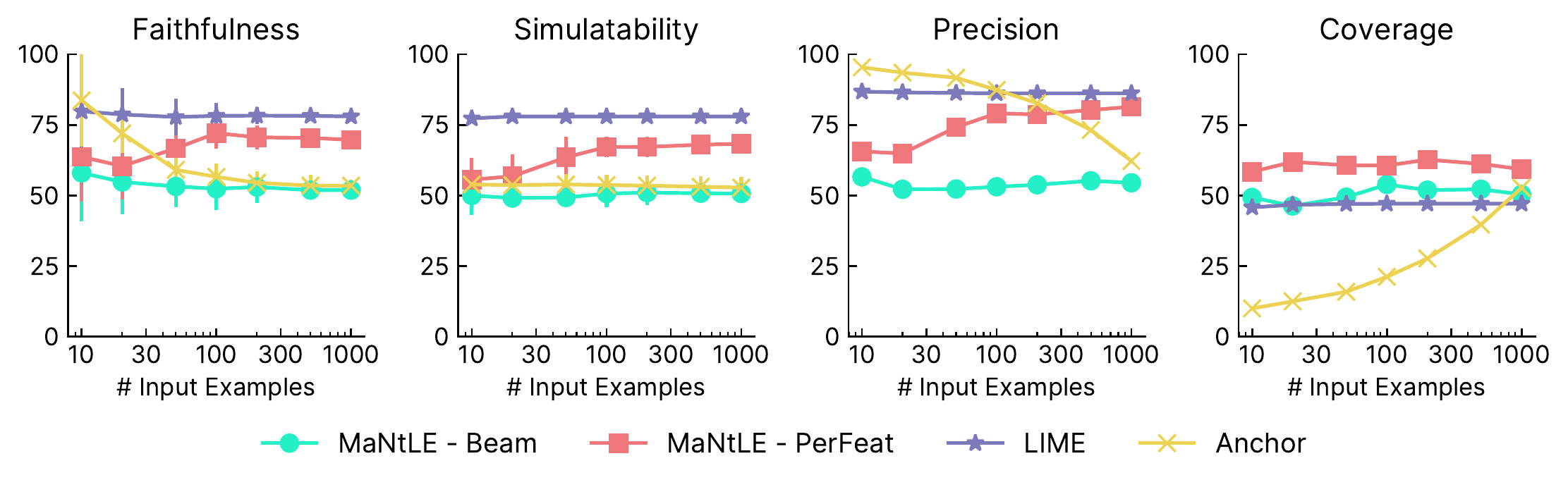}
         \caption{XGBoost Classifier}
         \label{fig:xgb_sampleexp}
     \end{subfigure}
    \caption{\textbf{Increasing number of input examples improves simulatability and faithfulness of \model-PF explanations.} Here, we show the results by increasing the number of input examples between 10 and 1000 for different classifiers trained on the Adult dataset. While the simultability of LIME, Anchor, and \model remain constant with an increase in the number of input examples, explanations from \model-PF improve simulatability and faithfulness.}
    \label{fig:app_sampleexp}
\end{figure*}

Extending on the results from \secref{sec:more_examples}, which presented the performance of {\tt logistic regression} and {\tt decision tree} models trained on the {\tt adult} dataset, we further investigate the performance of {\tt neural network} and {\tt xgboost} classifiers in this section. In addition, we evaluate the precision and coverage of the explanations on the simulation set (i.e., the test set).  Consistent with our findings in \secref{sec:more_examples}, increasing the number of examples used by \model consistently improves the precision of the explanations, leading to improved overall performance (Figure \ref{fig:app_sampleexp}).

\subsection{Can \model handle tasks with more features?}
\label{sec:more_features}
In this experiment, we evaluate the ability of \model, and the \Perfeat decoding variant of \model, to generate explanations for logistic regression and decision tree classifiers as we increase the number of features from 5 to 11 for the \texttt{adult} dataset. 
As in the prior sections, we measure the faithfulness and simulatability metrics.

In \secref{sec:interpret_classifiers}, we experimented with exactly five features for all datasets. 
However, in real-world situations, classifiers may operate over more than five features, which is why this evaluation is essential.

\begin{figure*}
    \centering
    \begin{subfigure}[b]{\textwidth}
         \centering
         \includegraphics[width=\textwidth]{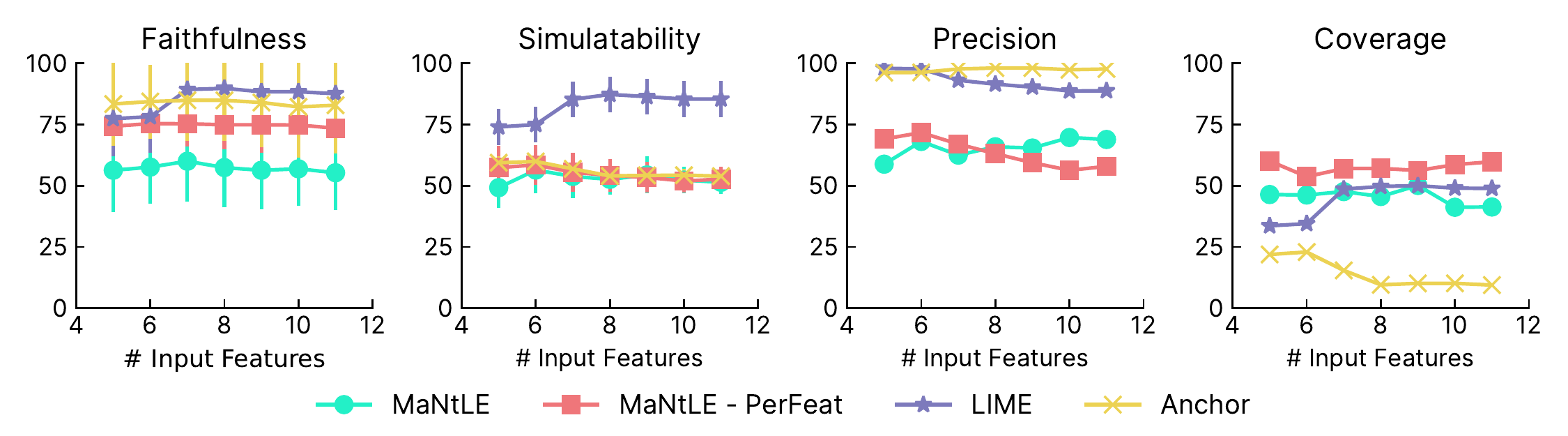}
         \caption{Logistic Regression}
         \label{fig:lr_featexp}
     \end{subfigure}
     \begin{subfigure}[b]{\textwidth}
         \centering
         \includegraphics[width=\textwidth]{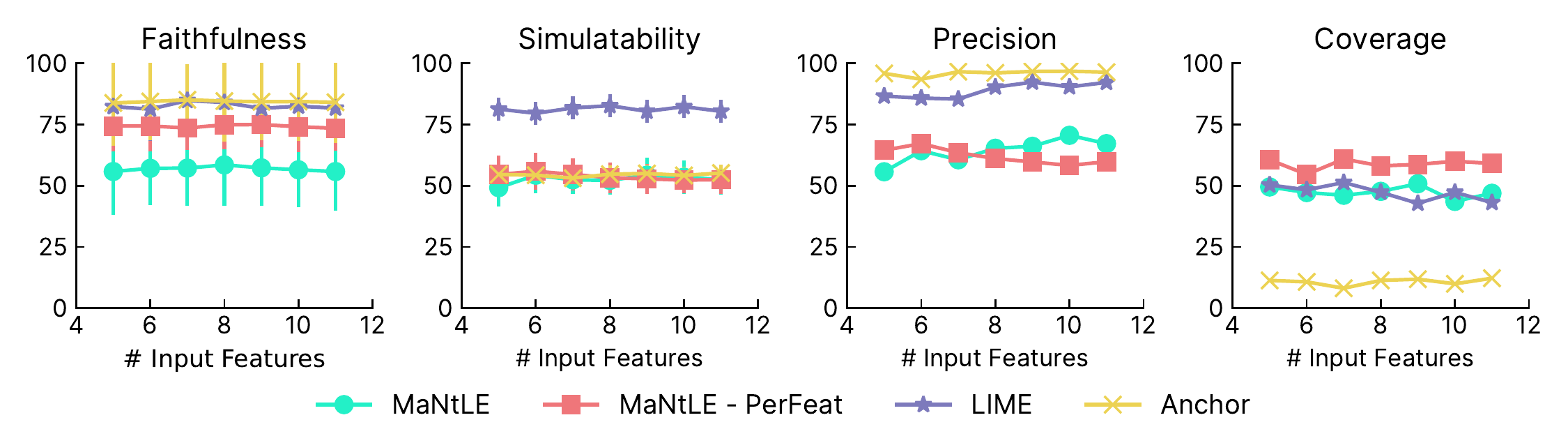}
         \caption{Decision Tree}
         \label{fig:dt_featexp}
     \end{subfigure}
     \begin{subfigure}[b]{\textwidth}
         \centering
         \includegraphics[width=\textwidth]{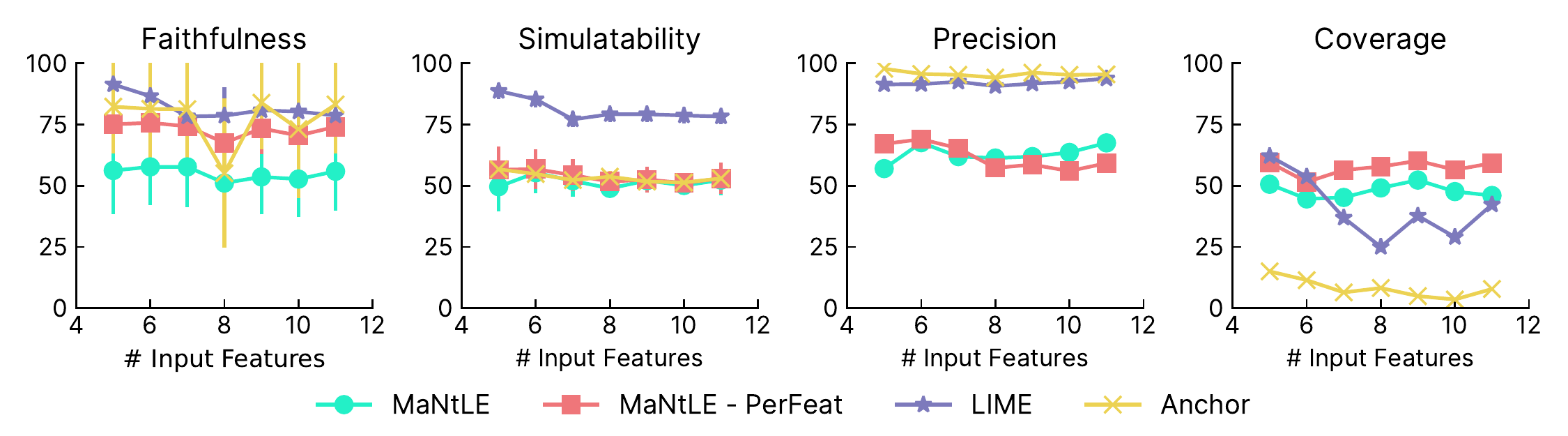}
         \caption{Neural Network}
         \label{fig:nn_featexp}
     \end{subfigure}
     \begin{subfigure}[b]{\textwidth}
         \centering
         \includegraphics[width=\textwidth]{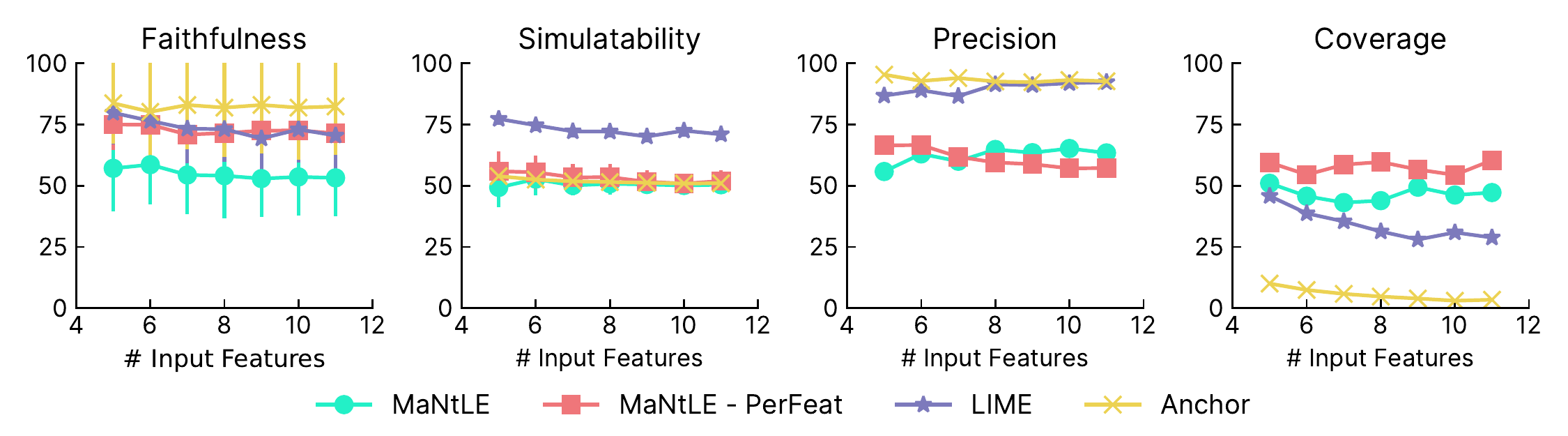}
         \caption{XGBoost}
         \label{fig:xgb_featexp}
     \end{subfigure}
    \caption{\textbf{Faithfulness of \model-PF explanations is largely independent of the number of input features across models}. Here, we compute metrics over 100 runs for the Adult dataset. The simulatability of explanations decreases with an increase in the number of features, as would be naturally expected.}
    \label{fig:app_featexp}
\end{figure*}

The results in Figure \ref{fig:app_featexp} suggest that the faithfulness of generated explanations is invariant to the number of input features used by a classifier that we seek to explain. 
However, while \model-PF generates more faithful explanations than \model, this advantage does not translate to improved simulatability as the number of features increases.

\section{Human Evaluation}
In this section, we expound on some nuances of our human evaluation and also provide templates as well as numeric results for the accuracies of people on solving the adult task using the different explanations.

Firstly, during a pilot study, we identified that workers generally had a 
hard time interpreting the meanings of different quantifiers and what confidence values to map such accuracies towards while making predictions on new examples. As a result, following the confidence values in \citet{srivastava-etal-2018-zero}, we reverse map the quantifiers in \model explanations to confidence values and present them to the turkers. An example of the conversion is \emph{`if Education not equal to Dropout, then Income is certainly >50K'} to \emph{`95\% of the time, the Income is >50K if Education not equal to Dropout'}. Secondly, we also noticed that workers had a preference for explanations that were of high confidence and often rated explanations as poor purely on the grounds that it wasn't of high confidence. Since comparison between baselines and \model would not be fair in such scenarios, we restrict human evaluations to settings where \model explanations are at least 85\% confident of their explanations (where confidence is measured by the quantifier used in the generated explanation). During experiments, we ask workers to simulate classification performance for the different classifiers used in our simulation experiments.

\begin{figure*}
    \centering
    \includegraphics[width=\textwidth]{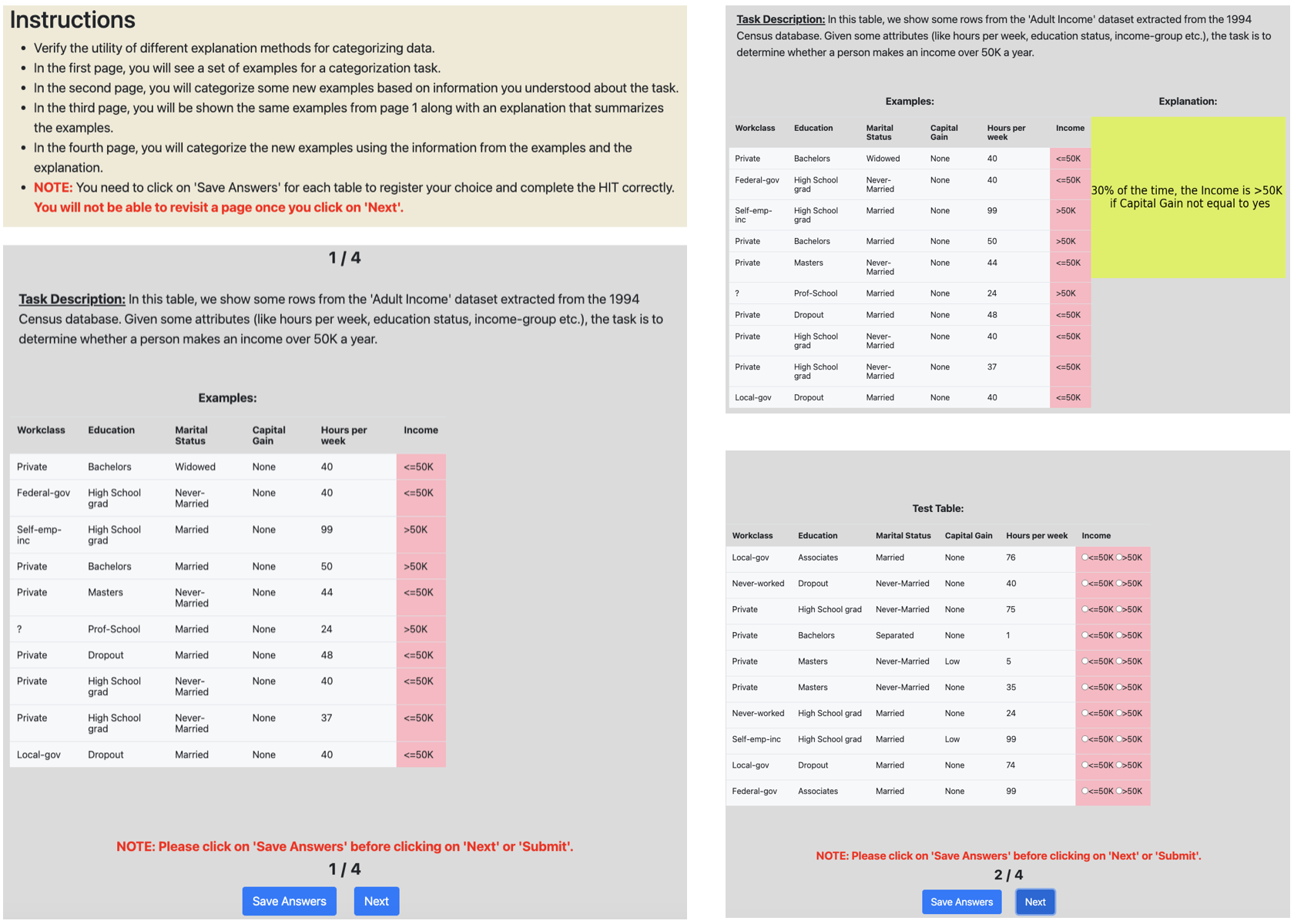}
    \caption{Template for performing the perturbation and simulation experiments using examples from the {\tt adult} dataset.}
    \label{fig:temp_classify}
\end{figure*}

\begin{figure*}
    \centering
    \includegraphics[width=\textwidth]{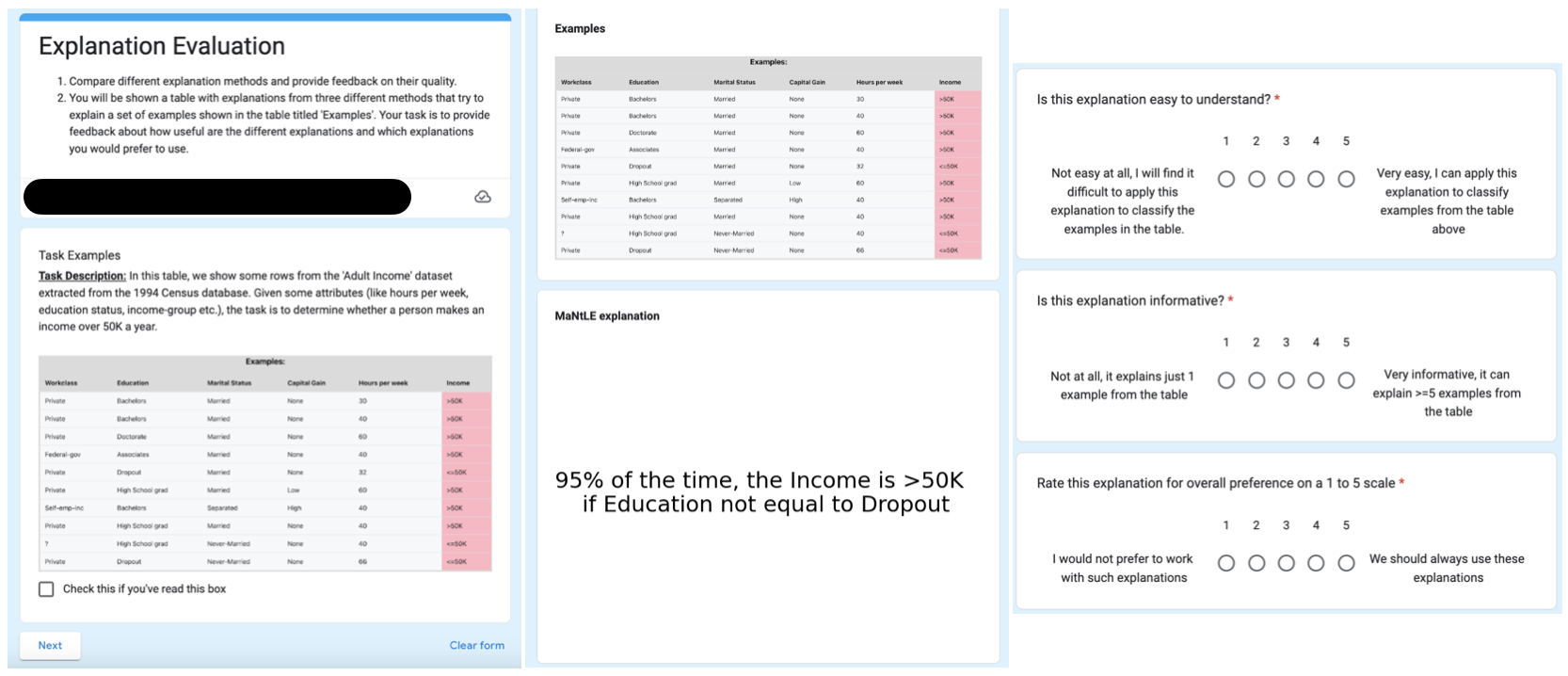}
    \caption{Template for performing the subjective evaluation of different explanations on a 1-5 Likert scale for understandability, informativeness, and overall preference.}
    \label{fig:temp_sub}
\end{figure*}

\begin{figure*}
    \centering
    \begin{subfigure}[b]{0.32\textwidth}
         \centering
         \includegraphics[width=\textwidth]{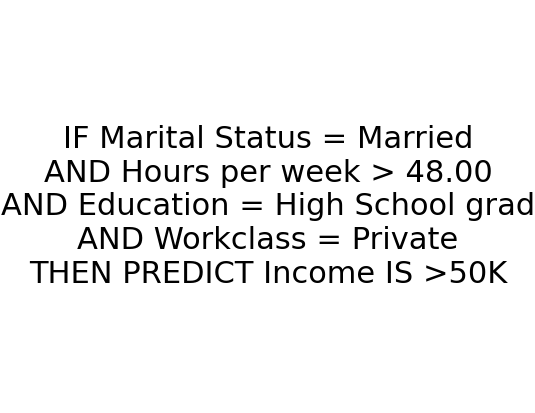}
         \caption{Anchors}
         \label{fig:expl_anchor}
     \end{subfigure}
     \begin{subfigure}[b]{0.32\textwidth}
         \centering
         \includegraphics[width=\textwidth]{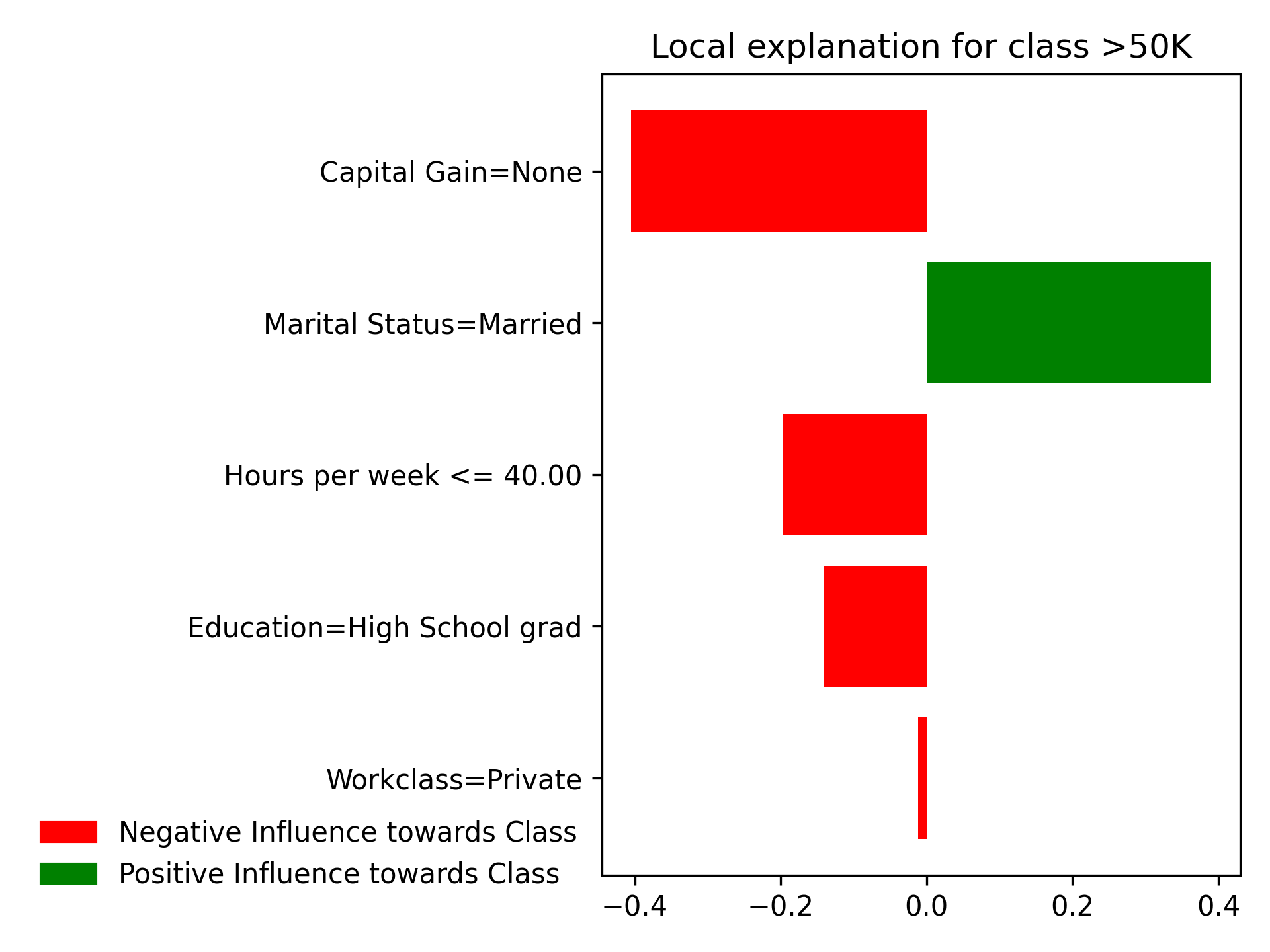}
         \caption{\lime}
         \label{fig:expl_lime}
     \end{subfigure}
     \begin{subfigure}[b]{0.32\textwidth}
         \centering
         \includegraphics[width=\textwidth]{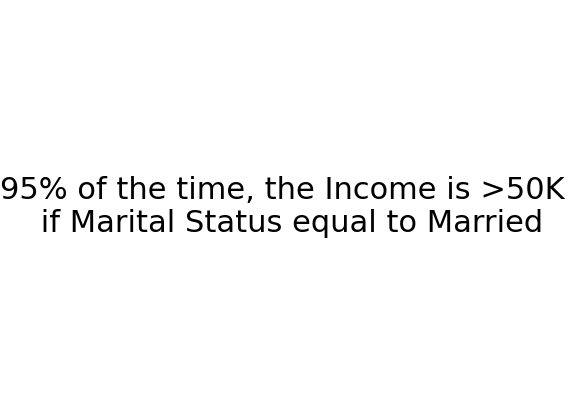}
         \caption{\model}
         \label{fig:expl_mantle}
     \end{subfigure}
    \caption{Presentation format of explanations used in human evaluation experiments.}
    \label{fig:expl_show}
\end{figure*}

The templates used in our experiments can be found in Figure \ref{fig:temp_classify} (for perturbation and simulation experiments) and Figure \ref{fig:temp_sub} (for explanation preference experiments). Examples of how the three explanations were presented to workers can be found in Figure \ref{fig:expl_show}.

Additionally, we present the classification accuracy (averaged over workers) during the pre-explanation and post-explanation phase of the perturbation and simulation experiments described in \secref{sec:human_exp}. 
It is worth noting that individual workers may have varying degrees of pre-explanation accuracy, thereby making a direct comparison of raw accuracies between explanation methods unfeasible. However, our primary objective is to investigate whether explanations improve the workers' classification ability. Therefore, we depict the percentage of instances where the explanations led to improved classification performance in Figure \ref{fig:human_eval}.\footnote{When we mention the classification performance of human workers, we refer to the number of times they can predict the same label as the classifier, whose examples and explanations they see during the learning phase.}

\begin{table}[ht!]
\normalsize
\begin{center}
\scalebox{0.94}{
\begin{tabular}{ll|r|r} 
 \toprule
 \textbf{Experiment} & \begin{tabular}{@{}l@{}}\textbf{Exp.}\\ \textbf{Method}\end{tabular} & \begin{tabular}{@{}l@{}}\textbf{Pre-Exp.}\\ \textbf{Accuracy}\end{tabular} & \begin{tabular}{@{}l@{}}\textbf{Post-Exp.}\\ \textbf{Accuracy}\end{tabular}\\
 \midrule
 & \lime & 71.5 & 67.7 (\textcolor{BrickRed}{↓}) \\
 Perturbation & \anchors & 63.1 & 55.4 (\textcolor{BrickRed}{↓}) \\
 & \model & 53.8 & 60.8 (\textcolor{ForestGreen}{\textbf{↑}}) \\
\midrule
 & \lime & 60.0 & 56.9 (\textcolor{BrickRed}{↓}) \\
 Simulation & \anchors & 67.7 & 63.8 (\textcolor{BrickRed}{↓}) \\
 & \model & 66.1 & 67.7 (\textcolor{ForestGreen}{\textbf{↑}}) \\
\bottomrule
 \end{tabular}
 }
 \end{center}
\vspace{-0.1in}
\caption{Average classification accuracies of workers on perturbation and simulation experiments for the {\tt adult} dataset in the pre-explanation and post-explanation phases. Overall, 23 workers took part in this study. Exp.= explanation}
\vspace{-0.1in}
\label{tab:human_acc_table}
\end{table}

\end{document}